\documentclass{article}

\usepackage[nonatbib, final]{neurips_2021}

\usepackage[utf8]{inputenc} 
\usepackage[T1]{fontenc}    
\usepackage{hyperref}       
\usepackage{url}            
\usepackage{booktabs}       
\usepackage{amsfonts}       
\usepackage{nicefrac}       
\usepackage{microtype}      
\usepackage{xcolor}         
\usepackage{graphicx}
\usepackage{subfigure}
\usepackage{amsmath}
\usepackage{multirow}
\usepackage{xfrac}
\usepackage[numbers]{natbib}

\newcommand*\numcircledmod[1]{\raisebox{.5pt}{\textcircled{\raisebox{-.9pt} {#1}}}}

\title{Hard-Attention for Scalable Image Classification}

\author{Athanasios Papadopoulos$^{1}$\quad Paweł Korus$^{1,2}$\quad Nasir Memon$^{1}$\\
$^1$Tandon School of Engineering, New York University \\
$^2$AGH University of Science and Technology \\
\texttt{\{tpapadop, pkorus, memon\}@nyu.edu} \\
}

\begin{document}

\maketitle

\begin{abstract}

Can we leverage high-resolution information without the unsustainable quadratic complexity to input scale?
We propose Traversal Network (TNet), a novel multi-scale hard-attention architecture, which traverses image scale-space in a top-down fashion, visiting only the most informative image regions along the way.
TNet offers an adjustable trade-off between accuracy and complexity, by changing the number of attended image locations.
We compare our model against hard-attention baselines on ImageNet, achieving higher accuracy with less resources (FLOPs, processing time and memory).
We further test our model on fMoW dataset, where we process satellite images of size up to $896 \times 896$ px, getting up to $2.5$x faster processing compared to baselines operating on the same resolution, while achieving higher accuracy as well.
TNet is modular, meaning that most classification models could be adopted as its backbone for feature extraction, making the reported performance gains orthogonal to benefits offered by existing optimized deep models.
Finally, hard-attention guarantees a degree of interpretability to our model's predictions, without any extra cost beyond inference.
Code is available at \url{https://github.com/Tpap/TNet}.

\end{abstract}

\section{Introduction}
\label{sec1}
In image classification, deep neural networks (DNNs) are typically designed and optimized for a specific input resolution, e.g. $224 \times 224$ px. Using modern DNNs on images of higher resolution (as happens e.g., in satellite or medical imaging) is a non-trivial problem due to the subtlety of scaling model architectures~\citep{tan2019efficientnet}, and rapid increase in computational and memory requirements.

A linear increase in the spatial dimensions of the input, results in a quadratic increase in computational complexity and memory, and can easily lead to resource bottlenecks.
This can be mitigated with careful engineering, e.g., streaming~\cite{pinckaers2019streaming} or gradient checkpointing~\cite{marra2020full}. However, such solutions are content-agnostic, and don't take advantage of the fact that discriminative information may be sparse and distributed across various image scales, deeming processing of the whole input unnecessary.

\begin{figure}[!t]
\begin{center}
\centerline{\includegraphics[width=\columnwidth]{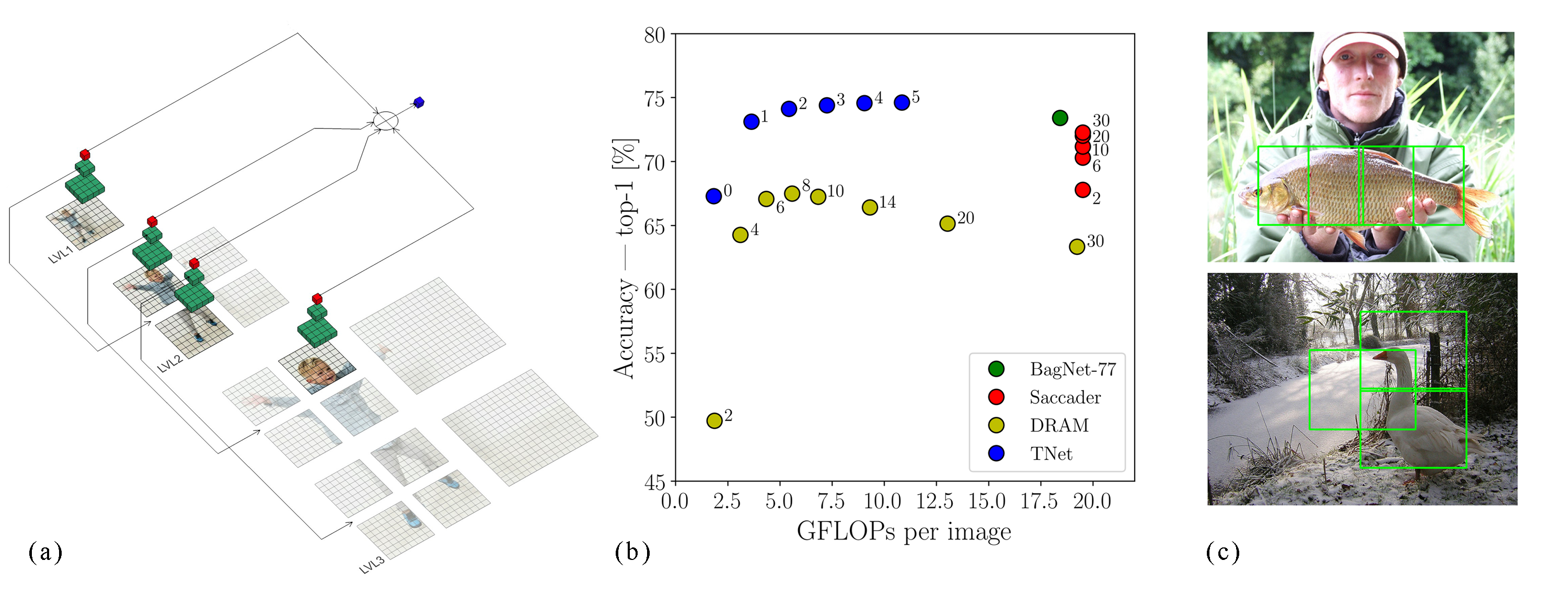}}
\caption{(a) Multi-scale processing in TNet. Starting at level $1$, features are selectively extracted from image regions at various scales (red cubes), and then, they are combined to create the final image representation used for classification (blue cube).
(b) Experimental results on ImageNet \cite{deng2009imagenet} with baselines based on \cite{elsayed2019saccader}. Numeric annotations correspond to the number of attended locations.
Our model offers a better trade-off between accuracy and complexity (FLOPs).
(c) Examples of attention policy (top $3$ locations) learned on ImageNet.}
\label{fig_1}
\end{center}
\vskip -0.2in
\end{figure}

Our goal is to leverage high-resolution information, while dispensing with the unsustainable quadratic complexity to input scale. To this end, we propose Traversal Network (TNet), a multi-scale hard-attention architecture, which traverses image scale-space in a top-down fashion, visiting only the most informative image regions along the way.
TNet is recursive, and can be applied to inputs of virtually any resolution; an outline of its processing flow is presented in Fig. \ref{fig_1} (a).
Our method draws its intuition from the way humans use saccades to explore the visual world.

TNet offers an adjustable trade-off between accuracy and complexity, by changing the number of attended image regions. This way, complexity increases linearly with the number of attended locations, irrespective of the input resolution.
Also, hard-attention explicitly reveals the image regions that our model values the most,
providing a certain degree of interpretability (Fig. \ref{fig_1} (c)).
Importantly, interpretability comes without any extra cost beyond inference, in contrast to popular attribution methods, which require at least an additional backward pass \cite{selvaraju2017grad}, or numerous forward passes \cite{ancona2017towards}.
Attention may also reduce data acquisition cost \cite{uzkent2020learning},
by allowing only a fraction of the high-resolution content to be acquired.

TNet is trained end-to-end by employing a modified version of REINFORCE rule \cite{williams1992simple}, while using only classification labels.
Our architecture is modular, and most classification models could be adopted as its backbone for feature extraction. This way, we can directly take advantage of various performance benefits offered by existing optimized deep models.

Hard-attention is the mechanism that allows TNet to dispense with quadratic complexity to input scale, and as a result, we evaluate our model against strong hard-attention baselines on ImageNet \cite{elsayed2019saccader}.
A summary of our results is depicted in Fig. \ref{fig_1} (b), where we see that TNet offers a better trade-off between accuracy and complexity measured in FLOPs (similar behavior is observed with actual timings and memory).
We extend our experiments to fMoW dataset, which consists of high-resolution satellite images \cite{fmow2018}. We process images up to $896 \times 896$ px, getting up to $2.5$x faster processing compared to baselines operating on the same resolution, while achieving higher accuracy as well.

We find improvements in accuracy surprising, because TNet is processing only part of the input, in contrast to fully convolutional baselines.
We primarily attribute this behavior to a novel regularization method, which encourages classification based on individual attended locations. We verify its efficacy through an ablation study.

\section{Related work}
\label{sec2}
\noindent \textbf{Attention.}
Attention has a long history in the artificial neural networks literature \cite{itti1998model}, and in the modern era of deep learning it has been used very successfully in various problems \cite{larochelle2010learning, denil2012learning, bahdanau2014neural, xu2015show, gregor2015draw, wang2018non, shen2020interpretable}. 
Two main forms of attention are:
\emph{soft attention} which processes everything but weights various regions differently; and \emph{hard attention} which selects only a fraction of the data for processing.
Hard-attention models address various use-cases, and can be motivated by interpretability~\cite{elsayed2019saccader}, reduction of high-resolution data acquisition cost \cite{uzkent2020learning}, or computational efficiency~\cite{katharopoulos2019processing}. Our goal is to offer a single model that takes advantage of all these benefits.

Our model is conceptually similar to glimpse-based models~\cite{ba2014multiple, mnih2014recurrent, ranzato2014learning, sermanet2014attention, eslami2016attend, ba2016using}.
An important difference is that we don't restrict our attention policy to $2$D space, but we consider scale dimension as well.
Also, our model parallelizes feature extraction at each processing level, instead of being fully sequential.
Furthermore, we don't use recurrent neural networks (RNNs) to combine
features from different locations, but instead, we simply average them (see Section \ref{sec3}). This way, gradients flow directly to extracted features, without the need to backpropagate through RNN steps.
This is a simpler strategy compared to LSTM in order to avoid vanishing gradients.

Recent work explores soft attention mechanisms based on transformers, which originate from the natural language processing community \cite{vaswani2017attention}.
Transformers have already been used extensively in machine vision \cite{parmar2018image, zhao2020exploring, dosovitskiy2020image, carion2020end, ramachandran2019stand}, and research interest in related directions has increased \cite{bello2019attention, chen20182, bello2021lambdanetworks}.

\noindent \textbf{Multi-scale representations.}
We identify four broad categories of multi-scale processing methods. $(1)$ \textit{Image pyramid methods} extract multi-scale features by processing multi-scale inputs \cite{eigen2014depth, pinheiro2014recurrent, ke2017multigrid, najibi2018autofocus}.
Our model belongs to this category, and due to its recursive nature, it can extract features from an arbitrary number of pyramid levels (see Section \ref{sec3}).
$(2)$ \textit{Encoding schemes} take advantage of the inherently hierarchical nature of deep neural nets, and reuse features from different layers, since they contain information of different scale \cite{he2014spatial, liu2016ssd, chen2018deeplab}.
$(3)$ \textit{Encoding-Decoding schemes} follow up the feed-forward processing (encoding) with a decoder, that gradually recovers the spatial resolution of early feature maps, by combining coarse and fine features \cite{ronneberger2015u, lin2017feature}.
$(4)$ \textit{Spatial modules} are incorporated into the forward pass, to alter feature extraction between layers \cite{yu2015multi, chen2017rethinking, wang2019elastic}.

\noindent \textbf{Computational efficiency.}
There are multiple ways to adjust the computational cost of deep neural networks. We organize them into four categories.
$(1)$ \textit{Compression methods} aim to remove redundancy from already trained models \cite{lecun1990optimal, hinton2015distilling, yu2018nisp}.
$(2)$ \textit{Lightweight design strategies} are used to replace network components with computationally lighter counterparts \cite{jaderberg2014speeding, iandola2016squeezenet, rastegari2016xnor, howard2017mobilenets, wang2017factorized}.
$(3)$ \textit{Partial computation methods} selectively utilize parts of a network, creating paths of computation with different costs \cite{larsson2016fractalnet, figurnov2017spatially, zamir2017feedback, huang2017multi, wu2018blockdrop}.
$(4)$ \textit{Attention methods} selectively process parts of the input, based on their importance for the task at hand \cite{ramapuram2018variational, levi2018efficient, katharopoulos2019processing, shen2021interpretable}.
This is the strategy we follow in our architecture.

\section{Architecture}
\label{sec3}
\begin{figure*}[!t]
\centering
\includegraphics[width=0.9\textwidth]{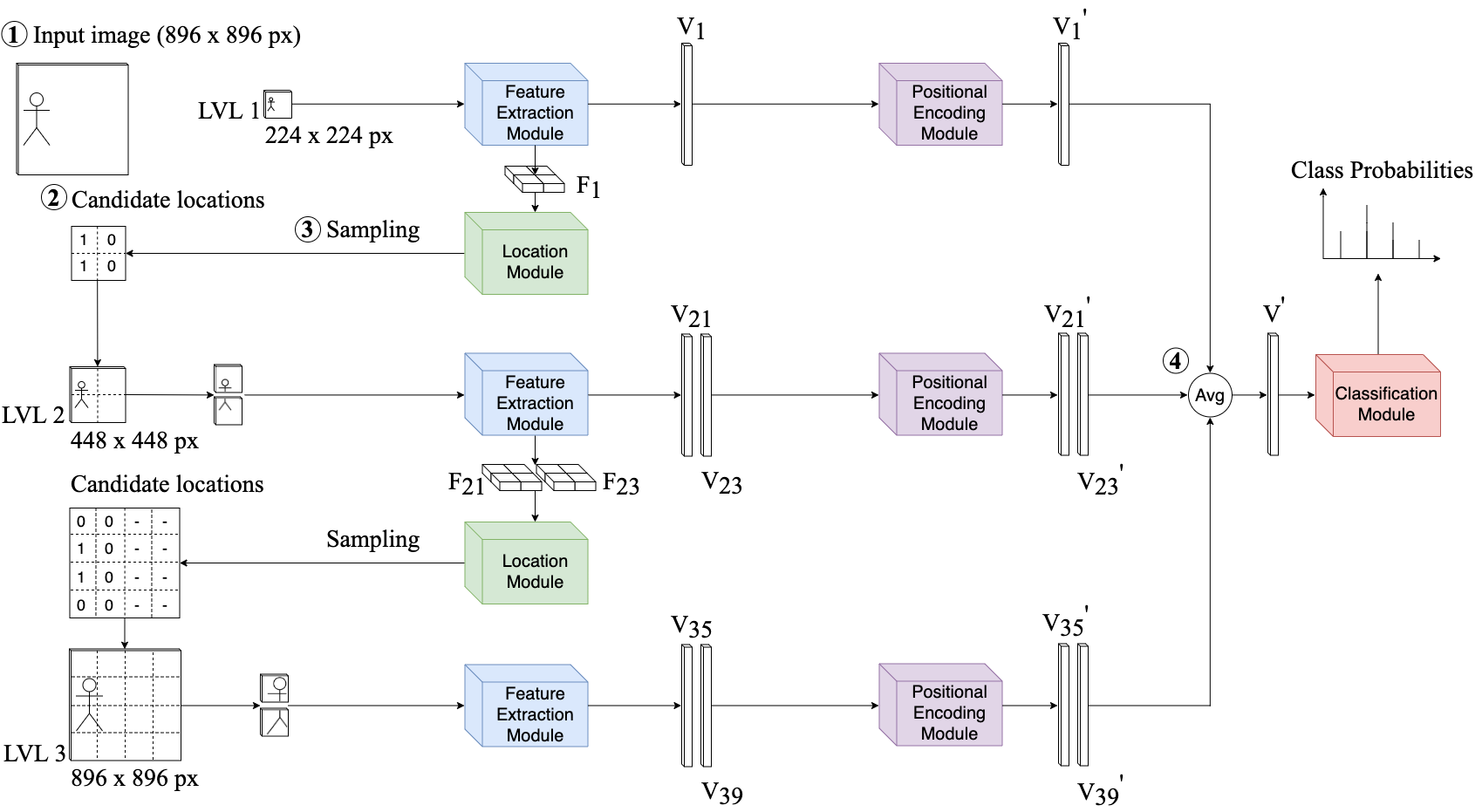}
\caption{Three unrolled processing levels of our architecture. Starting at level $1$, the image is processed in the coarsest scale (\emph{Feature Extraction Module}), and the extracted features are used to decide which image locations should be processed in finer detail (\emph{Location Module}). This process is repeated for each selected location to reach level $3$, where features from the highest resolution are extracted. All features are enriched with positional information
(\emph{Positional Encoding Module}), and then are averaged before the final classification (\emph{Classification Module}).}
\label{fig_4}
\end{figure*}

\subsection{Processing flow}
\label{sec3_1}
We present our architecture by walking through the example in Figure \ref{fig_4}, where we process an image with original resolution of $896 \times 896$ px (\numcircledmod{1} in the top left corner). 
In the first level, we downscale the image to $224 \times 224$ px and pass it through the \emph{feature extraction module}, in order to produce a feature vector $V_1$ that contains a coarse description of the original image.

To proceed to the next level, we feed an intermediate feature map, $F_{1}$, from the \emph{feature extraction module} to the \emph{location module}, which considers a number of candidate locations described by $F_1$, and predicts their importance (in this particular example, the candidate locations form a $2 \times 2$ regular grid (\numcircledmod{2}), and the \emph{location module} yields $4$ predictions).
We express region importance as attendance probability, which parametrizes a categorical distribution used for sampling without replacement; in our current example, we sample $2$ locations (\numcircledmod{3}).

In the 2nd processing level, we crop the selected regions from the full-resolution image, resize them to $224 \times 224$ px, and feed them to the \emph{feature extraction module} to obtain the corresponding feature vectors (here $V_{21}$ and $V_{23}$).
The original input resolution allows us to move to a $3$rd processing level, where we feed $F_{21}$ and $F_{23}$ to the \emph{location module}, leading to $2$ Categorical distributions. We sample $1$ location from each one of them to get $V_{35}$ and $V_{39}$. 

Features extracted from all levels are passed through the \emph{positional encoding module}, which injects information about the spatial position and scale of the image regions the features describe.
The resulting vectors, $\lbrace V^{'}_{*} \rbrace$, are averaged (\numcircledmod{4}) into a single comprehensive representation, $V^{'}$, that is fed to the \emph{classification module} for the final prediction.

\subsection{Modules}
\label{sec3_2}
The \textbf{feature extraction module} receives an image of fixed size as input, and outputs a feature vector $V$ and an intermediate spatial representation $F$. The input size, a hyperparameter we call \textit{base resolution}, defines the minimum amount of information that can be processed. Hence, it constrains the minimum cost that our model has to pay in computational and memory resources. In our experiments, feature extraction modules are implemented using CNNs.

The \textbf{location module} predicts $K=n^2$ probabilities of a Categorical distribution over the locations within a given $n \times n$ grid of candidate image regions.
It receives a feature map of size $n \times n \times c$ as input, where each $1 \times 1 \times c$ vector describes the corresponding location of the grid.
The feature map is passed through a series of $1 \times 1$ convolutions (contextual information is infused as well, e.g., via squeeze and excitation~\cite{hu2018squeeze}), yielding $K$ logits, which are transformed to relative region importance via a softmax layer.

The \textbf{positional encoding module} receives a feature vector $f$ and a positional encoding vector $p$, and combines them (e.g., through a fully connected layer) to an output feature vector $f^{'}$. We use a variant of the fixed positional encodings based on sine and cosine functions introduced in \cite{vaswani2017attention}. Instead of a single dimension of time, we have three:  two spatial dimensions and scale.

The \textbf{classification module} projects the final feature vector (e.g., via a linear layer) to classification logits. 
We provide the exact module architectures in Appendix \ref{App_sec_2_1}, along with justification of our design choices.

\section{Training}
\label{sec4}
\subsection{Learning rule}
\label{sec4_1}
Our model is not end-to-end differentiable because of location sampling. We address this problem using a variant of the \emph{REINFORCE} \cite{williams1992simple} learning rule:
\begin{equation}
\label{eq1}
L_F = \frac{1}{N \cdot M} \sum_{i=1}^{N \cdot M} \Big[ \frac{\partial \log{p(y_i|l^{i}, x_i, w)}}{\partial w} + \lambda_f (R_{i} - b) \frac{\partial \log{p(l^{i}|x_i, w)}}{\partial w} \Big]
\end{equation}

where $x_i$ is the $i$-th image, $y_i$ is its label, and $w$ are the parameters of our model.
$p(l^i|x_i, w)$ is the probability that the sequence of locations $l^i$ is attended for image $x_i$, and $p(y_i|l^i, x_i, w)$ is the probability of predicting the correct label after attending to $l^i$.

$N \cdot M$ is the total number of examples used for each update. The size of our original batch $B$ is $N$, and we derive \eqref{eq1} using a Monte Carlo estimator with $M$ samples to approximate the expectation $\sum_{l^i} p(l^i|x_i, w) \Big[ \frac{\partial \log{p(y_i|l^i, x_i, w)}}{\partial w} + \log{p(y_i|l^i, x_i, w)} \frac{\partial \log{p(l^i|x_i, w)}}{\partial w} \Big]$ for each image $x_i$ in $B$.
To reduce the variance of the estimator, we replace $\log{p(y_i|l^{i}, x_i, w)}$ with a discrete indicator function $R_i$, which is equal to $1$ for correct predictions and $0$ otherwise \cite{ba2014multiple}.
To the same end, we use baseline $b$, which corresponds to the exponential moving average of the mean reward $R_{i} \; \forall i$, and is updated after processing each training batch \cite{xu2015show}:
\begin{equation}
\label{b_eq}
b_n = 0.9 \cdot b_{n-1} + 0.1 \cdot \frac{1}{NM} \sum_{i=1}^{NM} R_i^n
\end{equation}

where $R_i^n$ is the reward for the $i$-th image in the $n$-th batch. For simplicity, we drop the subscript of $b_n$ in \eqref{eq1}. $\lambda_f$ is a weighting hyperparameter.
We provide a detailed derivation of our learning rule in Appendix \ref{App_sec1_1}.

The first term of $L_F$ is used to update the parameters in order to maximize the probability of the correct label. The second term is used to update the location selection process, according to the utility of the attended sequence $l^i$ in the prediction of the correct label.

\subsection{Per-feature regularization}
\label{sec4_2}
When our model attends to a location sequence $l_i$ and makes a correct prediction, we positively reinforce the probability to attend every location in the sequence.
This is expressed in the second term of \eqref{eq1}, where we use probability $p(l^{i}|x_i, w)$ of the whole sequence $l_i$.
However, some locations may not contribute to the correct prediction, e.g., if they have missed the object of interest.
In such cases, we reinforce the attendance of uninformative regions, encouraging a sub-optimal policy.

To mitigate this problem, for every attended location, we use its feature vector to make a separate classification prediction. Then, we use these predictions to complement our assessment on whether the corresponding image regions have useful content.
We modify our learning rule as follows:
\begin{subequations}
\label{eq2}
\begin{align}
\tag{\ref{eq2}}
L_r = &L_F^s + \frac{1}{|l^i|} \sum_{k=1}^{|l^i|}{L_F^{k}}, \\
\label{eq2a}
L_F^s = &\frac{1}{N \cdot M} \sum_{i=1}^{N \cdot M} \Big[ \lambda_c \frac{\partial \log{p(y_i|l^{i}, x_i, w)}}{\partial w} + \lambda_f \lambda_r (R^s_i - b) \frac{\partial \log{p(l^{i}|x_i, w)}}{\partial w} \Big], \\
\label{eq2b}
L_F^k = &\frac{1}{N \cdot M} \sum_{i=1}^{N \cdot M} \Big[ (1-\lambda_c) \frac{\partial \log{p(y_i|l^{i}_{k}, x_i, w)}}{\partial w} + \lambda_f (1 - \lambda_r) (R^k_i - b) \frac{\partial \log{p(l^{i}_{k}|x_i, w)}}{\partial w} \Big]
\end{align}
\end{subequations}

where $L_F^{s}$ is learning rule \eqref{eq1} with additional weighting hyperparameters $\lambda_c, \lambda_r \in [0, 1]$.
$L_F^{k}$ is learning rule \eqref{eq1} when we attend only to the $k$-th location, $l^{i}_{k}$, from every sequence $l^{i}$. Also, we introduce weighting factors $(1 -\lambda_c)$ and $(1-\lambda_r)$.
During training, we attend to a fixed number of $|l^{i}|$ locations for every image.
$R^s_i$ and $R^k_i$ are discrete indicator functions equal to $1$ when a prediction is correct, and $0$ otherwise.
$L_F^{s}$ updates the parameters of our model based on attending to $l^i \; \forall i$, while $L_F^{k}$ updates them based on $l^{i}_{k}$; $\lambda_c$ and $\lambda_r$ specify the relative importance of these updates.

Even though our initial motivation was to improve the attention policy, the first term in \eqref{eq2b} updates feature extraction parameters based on independent predictions from attended image regions. We empirically observed that such updates boost performance, potentially because they lead to features that co-adapt less and generalize better.

\section{Experimental evaluation}
\label{sec5}

\subsection{Effectiveness of hard-attention mechanism}
\label{sec5_1}
\begin{table*}[!t]
\caption{Efficacy of hard-attention mechanism: TNet surpasses all hard-attention baselines \cite{elsayed2019saccader} on ImageNet, by attending to just $1$ location. Higher accuracy can be achieved for less FLOPs, and translates to lower actual run time. Memory savings are obtained compared to BagNet-$77$ and Saccader as well.
TNet has slightly more parameters compared to BagNet-$77$ due to the additional modules, but has significantly fewer parameters than the hard-attention baselines.
FLOPs, run time, and memory are measured during inference.
}
\label{table_1}
\begin{center}
\begin{small}
\begin{tabular}{lccccccc}
\toprule
\multirow{2}{*}{\textbf{Model}} & \multirow{2}{*}{\textbf{\#Locs}} & \multirow{2}{*}{\shortstack{\textbf{Top-$\mathbf{1}$} \\ \textbf{Acc.}}}  & \multirow{2}{*}{\shortstack{\textbf{Top-$\mathbf{5}$} \\ \textbf{Acc.}}} & \multirow{2}{*}{\shortstack{\textbf{FLOPs} \\ \textbf{(B)}}} & \multirow{2}{*}{\shortstack{\textbf{\#Params} \\ \textbf{(M)}}} & \multirow{2}{*}{\shortstack{\textbf{Time} \\ \textbf{(msec/im)}}} & \multirow{2}{*}{\shortstack{\textbf{Memory} \\ \textbf{(GB)}}} \\
& & & & & & & \\
\midrule
\multirow{4}{*}{\textbf{Saccader}} & $2$ & $67.79 \%$ & $85.42 \%$ & $19.5$ & \multirow{4}{*}{$35.58$} & $7.31 \pm 1.55$ & $2.58$ \\
& $6$ & $70.31 \%$ & $87.8 \%$ & $19.5$ & & $7.32 \pm 1.55$ & $2.52$ \\
& $20$ & $72.02 \%$ & $89.51 \%$ & $19.51$ & & $7.36 \pm 1.50$ & $2.53$ \\
& $30$ & $72.27 \%$ & $89.79 \%$ & $19.51$ & & $7.36 \pm 1.49$ & $2.51$ \\
\midrule
\multirow{4}{*}{\textbf{DRAM}} & $2$ & $49.72 \%$ & $73.27 \%$ & $1.86$ & \multirow{4}{*}{$45.61$} & $3.43 \pm 1.57$ & $0.45$ \\
& $4$ & $64.26 \%$ & $84.84 \%$ & $3.1$ & & $3.92 \pm 1.56$ & $0.44$ \\
& $8$ & $67.5 \%$ & $86.6 \%$ & $5.58$ & & $4.61 \pm 1.59$ & $0.45$ \\
& $20$ & $65.15 \%$ & $84.58 \%$ & $13.03$ & & $7.58 \pm 1.53$ & $0.46$ \\
\midrule
\textbf{BagNet-}$\mathbf{77}$ & - & $73.42 \%$ & $91.1 \%$ & $18.42$ & $20.55$ & $5.94 \pm 0.09$ & $2.62$ \\
\midrule
\multirow{4}{*}{\textbf{TNet}} & $0$ & $67.29 \%$ & $87.38 \%$ & $1.82$ & \multirow{4}{*}{$21.86$} & $0.74 \pm 0.01$ & $0.46$ \\
& $1$ & $73.12 \%$ & $90.56 \%$ & $3.63$ & & $1.43 \pm 0.01$ & $0.57$ \\
& $2$ & $74.12 \%$ & $91.18 \%$ & $5.43$ & & $2.09 \pm 0.02$ & $0.69$ \\
& $3$ & $74.41 \%$ & $91.4 \%$ & $7.24$ & & $2.74 \pm 0.03$ & $0.95$ \\
& $5$ & $74.62 \%$ & $91.35 \%$ & $10.84$ & & $3.96 \pm 0.04$ & $1.47$\\
\bottomrule
\end{tabular}
\end{small}
\end{center}
\vskip -0.1in
\end{table*}

\textbf{Data.}
ImageNet \cite{deng2009imagenet} consists of natural images from $1,000$ classes.
We use the ILSVRC $2012$ version, which consists of $1,281,167$ training and $50,000$ validation images.

\textbf{Models.} We use Saccader and DRAM \cite{elsayed2019saccader} as hard-attention baselines.
We use Saccader with BagNet-$77$-lowD \cite{brendel2019approximating} as its backbone for feature extraction. BagNet-$77$-lowD is based on ResNet-$50$ \cite{he2016deep}, with receptive field constrained to $77 \times 77$ px. DRAM uses the standard ResNet-$50$ for feature extraction, with glimpses of $77 \times 77$ px.

For fair comparison, we set the base resolution of TNet to $77 \times 77$ px, and use BagNet-$77$, a slightly modified version of BagNet-$77$-lowD, as the feature extraction module.
In the location module, we use a uniform $5 \times 5$ grid of overlapping candidate regions.
The dimensions of each grid cell span $34.375 \%$ of the corresponding image dimensions. This way, for an image of size $224 \times 224$ px, the image patches within the grid cells at the $2$nd processing level are $77 \times 77$ px.
We use BagNet-$77$ as a separate fully convolutional baseline, with inputs of $224 \times 224$ px.
Additional details about TNet and BagNet-$77$ are provided in Appendix \ref{App_sec_Arch_1}.

\textbf{Training.}
We train TNet with $2$ processing levels on images of $224 \times 224$ px using class labels only. We train for $200$ epochs using the Adam optimizer \cite{kingma2014adam} with initial learning rate $10^{-4}$, that we drop once by a factor of $0.1$. We use dropout (keep probability $0.5$) in the last layer of feature extraction. We use per-feature regularization with $\lambda_c=\lambda_r=0.3$. We attend to a fixed number of $3$ locations.

We train BagNet-$77$ from scratch, on $224 \times 224$ px images. Compared to TNet, we reduce dropout keep probability to $0.375$, and we early-stop at $175$ epochs.
We don't train our own Saccader and DRAM, we use results reported in \cite{elsayed2019saccader}.
Additional training details are provided in Appendix \ref{App_sec_2_2_1}.

\textbf{Results.} We present our results in Table \ref{table_1}. TNet outperforms Saccader and DRAM by attending to only $1$ location, while it surpasses BagNet-$77$ by attending to $2$. Saccader was designed with accuracy and interpretability in mind, which leads to sub-optimal computational efficiency, as it processes the entire full-resolution image before attending to locations. As a result, FLOPs stay nearly constant and remain similar to BagNet-$77$, more that $5$ times higher than TNet with $1$ attended location. DRAM offers the expected gradual increase in computational cost as the number of attended locations increases, but for FLOPs comparable with TNet - e.g., DRAM for $8$ locations (maximum accuracy) and TNet for $2$ - our model is superior by more than $6.5 \%$.

TNet has slightly more parameters than BagNet-$77$, because of the location and positional encoding modules. Saccader and DRAM have significantly heavier attention mechanisms in terms of parameters.

We profile all models to validate correspondence between theoretical FLOPs and real processing speed, and to assess memory requirements.
We time inference on batches of $64$ images; early batches are discarded to discount code optimization. We use a single NVIDIA Quadro RTX $8000$ GPU, with $64$ GB of RAM, and $20$ CPUs to mitigate data pipeline impact. For Saccader and DRAM we use public implementations \cite{saccader2021github} in TensorFlow (TF) $1$. TNet and BagNet-$77$ are implemented in TF $2$.
The difference in TF versions may be a confounding factor in the obtained results. However, TNet and BagNet-$77$ use the same environment and yield the expected difference in latency.
DRAM attends to locations sequentially, while TNet processes all locations from the same level in parallel. This leads to fixed memory use in DRAM and monotonic increase in TNet, as the number of attended locations increases. We could trade this for latency by switching to fully sequential processing.

\subsection{Scalability}
\label{sec5_2}
\begin{table*}[!t]
\caption{TNet effectively scales to images of resolution $448 \times 448$ px and $896 \times 896$ px on fMoW~\cite{fmow2018}. It surpasses in accuracy EfficientNet-B$0$ baselines trained on inputs of the same resolution, while it requires less FLOPs.
Differences in FLOPs translate to differences in actual run time, while memory requirements are lower as well.
TNet has more parameters compared to EficientNet-B$0$ due to the additional modules.
FLOPs, run time, and memory are measured during inference.
A graphical representation of the main results is provided in Appendix \ref{App_sec_4_1}.
}
\label{table_2}
\begin{center}
\resizebox{1.0\textwidth}{!}{
\begin{small}
\begin{tabular}{lccccccccc}
\toprule
\multirow{2}{*}{\textbf{Model}} & \multirow{2}{*}{\shortstack{\textbf{Input} \\ \textbf{Size}}} & \multirow{2}{*}{\textbf{\#Locs}} & \multirow{2}{*}{\textbf{BBoxes}} & \multirow{2}{*}{\shortstack{\textbf{Top-$\mathbf{1}$} \\ \textbf{Acc.}}}  & \multirow{2}{*}{\shortstack{\textbf{Top-$\mathbf{5}$} \\ \textbf{Acc.}}} & \multirow{2}{*}{\shortstack{\textbf{FLOPs} \\ \textbf{(B)}}} & \multirow{2}{*}{\shortstack{\textbf{\#Params} \\ \textbf{(M)}}} & \multirow{2}{*}{\shortstack{\textbf{Time} \\ \textbf{(msec/im)}}} & \multirow{2}{*}{\shortstack{\textbf{Memory} \\ \textbf{(GB)}}} \\
& & & & & & & & & \\
\midrule
\textbf{EfficientNet-B}$\mathbf{0}$ & $224^2$ & - & $\surd$ & $69.7 \%$ & $89.22 \%$ & $0.39$ & $4.13$ & $0.80 \pm 0.01$ & $0.76$ \\
\textbf{ResNet-}$\mathbf{50}$ \cite{uzkent2020learning} & $224^2$ & - & $\surd$ & $67.3 \%$ & - & $4.09$ & $23.71$ & - & - \\
\textbf{DenseNet-}$\mathbf{121}$ \cite{uzkent2019learning} & $224^2$ & - & $\surd$ & $70.7 \%$ & - & $3$ & $7.1$ & - & - \\
\midrule
\multirow{3}{*}{\textbf{EfficientNet-B}$\mathbf{0}$} & $224^2$ & \multirow{3}{*}{-} & \multirow{3}{*}{-} & $62.8 \%$ & $84.97 \%$ & $0.39$ & \multirow{3}{*}{$4.13$} & $0.80 \pm 0.01$ & $0.76$ \\
& $448^2$ & & & $69.83 \%$ & $90.22 \%$ & $1.54$ & & $3.04 \pm 0.18$ & $2.6$ \\
& $896^2$ & & & $70.6 \%$ & $90.81 \%$ & $6.18$ & & $12.18 \pm 0.18$ & $10.35$ \\
\midrule
\multirow{6}{*}{\textbf{TNet}} & $224^2$ & $0$ & - & $47.79 \%$ & $81.14 \%$ & $0.39$ & \multirow{6}{*}{$4.56$} & $0.84 \pm 0.02$ & $0.76$ \\
\cmidrule{2-10}
& \multirow{3}{*}{$448^2$} & $1$ & \multirow{3}{*}{-} & $70.17 \%$ & $90.99 \%$ & $0.77$ & & $1.74 \pm 0.02$ & $1.3$ \\
& & $2$ & & $71.46 \%$ & $91.58 \%$ & $1.16$ & & $2.46 \pm 0.03$ & $1.55$ \\
& & $3$ & & $71.57 \%$ & $91.72 \%$ & $1.55$ & & $3.18 \pm 0.04$ & $2.03$ \\
\cmidrule{2-10}
& \multirow{2}{*}{$896^2$} & $4$ & \multirow{2}{*}{-} & $72.16 \%$ & $91.98 \%$ & $1.94$ & & $4.71 \pm 0.03$ & $3.13$ \\
& & $6$ & & $71.92 \%$ & $91.83 \%$ & $2.71$ & & $6.13 \pm 0.05$ & $3.47$ \\
\bottomrule
\end{tabular}
\end{small}
}
\end{center}
\vskip -0.1in
\end{table*}

\textbf{Data.}
Functional Map of the World (fMoW) \cite{fmow2018} consists of high-resolution satellite images from $62$ classes. They are split in $363,572$ training, $53,041$ validation and $53,473$ testing images. All images have bounding box annotations.

\textbf{Models.}
We use EfficientNet-B$0$ (EN-B$0$) \cite{tan2019efficientnet} as the feature extraction module of TNet, with base resolution of $224 \times 224$ px. In the location module we use a $3 \times 3$ grid with overlapping cells ($50 \%$ overlap).
Hence, when processing $2$ levels, we use inputs of $448 \times 448$ px, and with $3$ levels, we use inputs of $896 \times 896$ px. Details of the TNet architecture are provided in Appendix \ref{App_sec_Arch_2}.

We create our first $3$ baselines by training EN-B$0$ with images resized to different resolution; $224 \times 224$ px, $448 \times 448$ px, and $896 \times 896$ px.
We use the available bounding boxes to crop the regions of interest, resize them to $224 \times 224$ px, and train an additional EN-B$0$ baseline.
Also, we report the best accuracy we find in the literature, achieved by single-model predictions (not network ensembles), and without the use of any meta-data for training. These are the ResNet-$50$~\cite{uzkent2020learning} and DenseNet-$121$~\cite{uzkent2019learning} baselines.

\textbf{Training.} We train TNet using only classification labels. The training proceeds in $2$ stages: (1) training from scratch for $40$ epochs on $448 \times 448$ px inputs, using $2$ processing levels, and a fixed number of $2$ locations; (2) fine-tuning for $10$ more epochs on images of $896 \times 896$ px, which allow us to extend processing to $3$ levels. We use $4$ locations; $2$ in the $2$nd processing level, plus $1$ for each of them in the $3$rd. We use per-feature regularization in both stages. All $4$ EN-B$0$ baselines are trained independently, from scratch. The exact hyperparameters are provided in Appendix \ref{App_sec_2_2_2}.

\textbf{Results.} We present our results in Table \ref{table_2}.
TNet surpasses all baselines by attending to $2$ locations ($2$ processing levels), with top-$1$ accuracy of $71.46 \%$. This requires less FLOPs than the corresponding EN-B$0$ baseline at the same input size ($448 \times 448$ px), which achieves accuracy of $69.83 \%$.

Extending processing to $3$ levels with $4$ attended locations, further increases accuracy to $72.16 \%$, while FLOPs increase by $\sim 67 \%$ (from $1.16$ to $1.94$B FLOPs).
The corresponding EN-B$0$ baseline operating on $896 \times 896$ px inputs, achieves $70.6 \%$ accuracy, and requires $\sim 300 \%$ more FLOPs compared to EN-B$0$ operating on $448 \times 448$ px inputs (from $1.54$ to $6.18$B FLOPs).
This shows that TNet can efficiently leverage high-resolution information.

We profile our models by following the procedure described in Section \ref{sec5_1}, and we show that differences in FLOPs translate to differences in actual run time. TNet also requires less memory, although it has more parameters because of the location and positional encoding modules.

We note
that when TNet limits its processing to just $1$ level (it uses only the contextual feature vector, $f_c$, extracted from the downscaled version of the whole image), its accuracy is considerably lower compared to EN-B$0$ operating on $224 \times 224$ px inputs.
We hypothesize that as TNet is trained with more processing levels, $f_c$ co-adapts with the increasing number of extracted feature vectors.
To mitigate this and maintain high accuracy at every processing level, as part of our future work, we can update Equation \ref{eq2b} by appropriately weighting terms that correspond to feature vectors from different levels.

\subsection{Modularity and fine-tuning}
\label{secMFT}
\begin{figure*}[!t]
\centering
\includegraphics[width=0.9\textwidth]{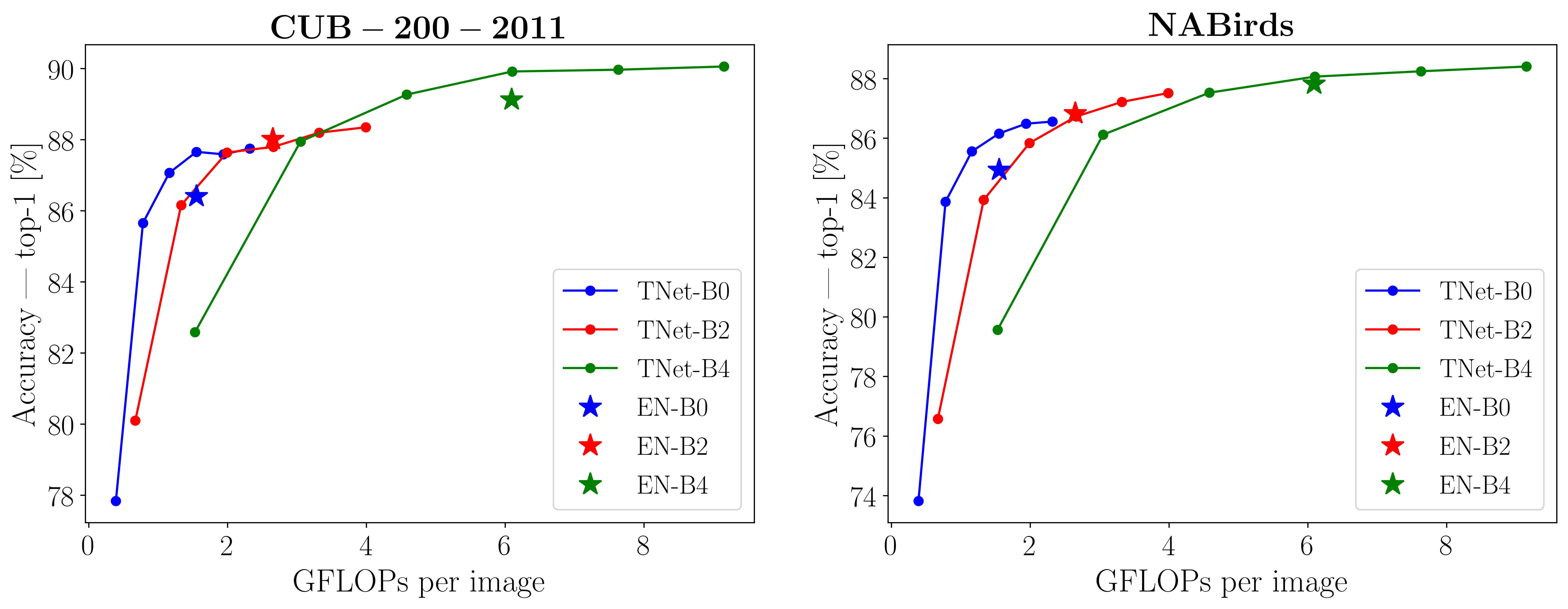}
\caption{TNet is modular, and different models can be used as its feature extraction module.
We plot TNet performance with $3$ different backbones (attended locations range from $0$ to $5$).
Stronger backbones lead to the expected increase in accuracy and FLOPs.
EN baselines and the feature extraction modules of TNet models are initialized with pre-trained weights, while the rest of the modules are randomly initialized.
TNet models achieve similar or better accuracy compared to corresponding baselines, showing that fine-tuning can be an effective practice.
}
\label{fig_5}
\end{figure*}

TNet is a modular architecture, and we would like to test its compatibility with different backbone models, especially when these models are initialized with pre-trained weights, since this is a popular practice \cite{devlin2018bert, dosovitskiy2020image, tan2021efficientnetv2}.

\textbf{Data.}
We use $2$ fine-grained classification datasets, CUB-$200$-$2011$ \cite{wah2011caltech} and NAbirds \cite{van2015building}, which are commonly used as downstream tasks.
CUB-$200$-$2011$ has $200$ classes of bird species, with $5,994$ training and $5,794$ testing images.
NABirds has $555$ classes of bird species, with $23,929$ training and $24,633$ testing images. 
We resize all images to $448 \times 448$ px.

\textbf{Models.} We use baselines from the EfficientNet (EN) family \cite{tan2019efficientnet}, EN-B$i$, $i \in \{0, 1, ...,4\}$.
We use the same EN models as the feature extraction module of TNet, getting TNet-B$i$.
All TNet models have base resolution of $224 \times 224$ px, attention grid of $5 \times 5$, and each grid cell dimension occupies $35\%$ of the corresponding image dimension. Processing extends to $2$ levels.

For TNet models, we use a weighted average to replace the simple averaging of the extracted feature vectors before the final prediction (\numcircledmod{4} in Fig. \ref{fig_4}).
The weights for the averaging are calculated by a new \emph{feature weighting module}, which receives the $N$ feature vectors extracted from all processing levels, and outputs $N$ weights that sum up to $1$.
Details about all architectures are provided in Appendix \ref{App_sec_Arch_3}.

\textbf{Training.} EN baselines and the feature extraction modules of TNet models are initialized with weights pre-trained on ImageNet \cite{efficientnet2021github}. The rest of TNet modules are randomly initialized. All models are fine-tuned on the downstream tasks. Training details are provided in Appendix \ref{App_sec_2_2_3}.

\textbf{Results.}
We summarize our results in Figure \ref{fig_5}. For clarity, we limit the number of models we plot,
and we provide detailed results in Appendix \ref{App_sec_4_2}.
We see that performance differences between baselines, translate to similar differences between TNet models with corresponding backbones, e.g., EN-B$4$ and TNet-B$4$ achieve considerably better accuracy compared to EN-B$0$ and TNet-B$0$ respectively.
This indicates that behavioral differences between classification networks can be manifested in TNet models as well, if these networks are used as feature extraction modules.

In addition, as attended locations increase, TNet models achieve similar or better accuracy compared to corresponding baselines (TNet-B$4$ is also competitive to strong baseline API-Net \cite{zhuang2020learning} and state-of-the-art TransFG \cite{he2021transfg}; see Appendix \ref{App_sec_4_2}).
This indicates that initialization with pre-trained weights allows TNet to learn useful weights for all modules.

\subsection{Attention policy and interpretability}
\label{sec5_3}
We show examples of attended locations in Fig. \ref{fig_1} (c) and Fig. \ref{fig_6}; more examples from all datasets are provided in Appendix \ref{App_sec_att_2}.
On ImageNet, the learned policy is highly content dependent and informative, as it overlaps with intuitively meaningful regions.
The same behavior is observed in CUB-$200$-$2011$ and NABirds datasets, where the weights estimated by the \emph{feature weighting module} have significant contribution to the interpretability of the predictions.
On fMoW, our model predominantly attends to locations at the center, where nearly all objects of interest are located.
While this is a correct heuristic implied by the data, it leads to a biased attention policy that is not sensitive to content changes. This demonstrates that sufficient diversity is needed to train fully informative policies.

To examine the relative importance between the coarse context and the information from the attended locations, we evaluate TNet on ImageNet, by using features only from attended locations. With $2$, $3$ and $5$ attended locations, TNet achieves top-$1$ accuracy $67.95 \%$, $69.66 \%$ and $71.05 \%$ respectively (with contextual information included, the corresponding accuracy values are $74.12 \%$, $74.41 \%$ and $74.62 \%$). The Saccader achieves similar accuracy for $2$, $4$ and $8$ locations, with $67.8 \%$, $69.51 \%$ and $70.08 \%$.
This shows that features extracted from attended locations have discriminative value even without global context. As a result, the attention policy can be valuable for interpreting predictions.

Hard-attention can also be used for reduction of high-resolution data acquisition cost \cite{uzkent2020learning}.
On fMoW, when our model attends to $4$ locations in $3$ processing levels, it covers $37.91 \%$ of the input in resolution of $448 \times 448$ px, and less than $12.5 \%$ in the highest resolution of $896 \times 896$ px.
Further quantitative analysis of the attention policy is provided in Appendix \ref{App_sec_att_1}.
\begin{figure*}[!t]
\centering
\includegraphics[width=1.0\textwidth]{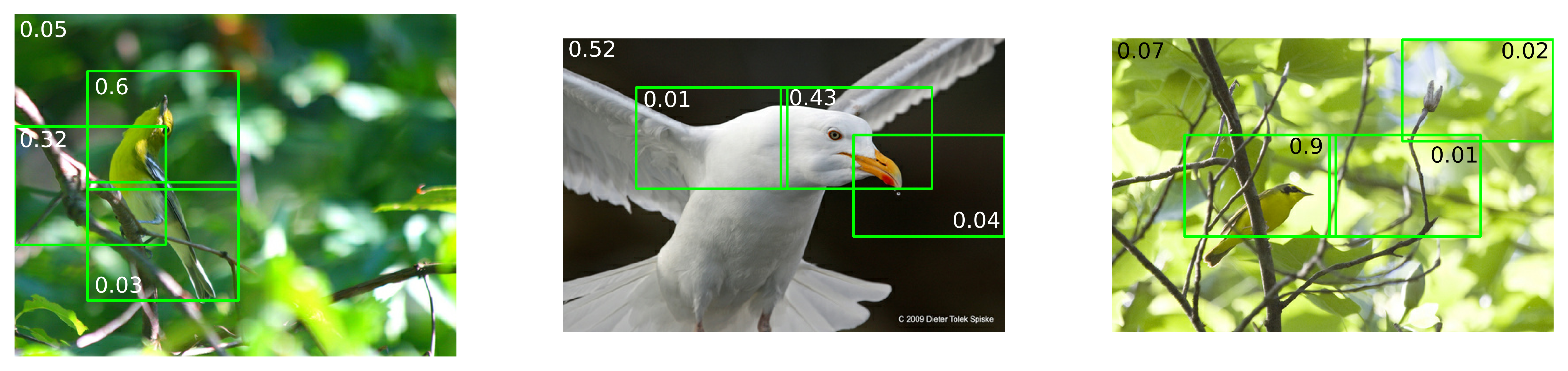}
\caption{Examples of attention policy learned on CUB-$200$-$2011$. Numeric annotations correspond to weights predicted by the \emph{feature weighting module} for the top $3$ locations and the downscaled version of the whole image ($1$st processing level). Weights sum up to $1$.
}
\label{fig_6}
\end{figure*}

\subsection{Ablation study}
\label{sec5_4}
We examine the effect of per-feature regularization by training TNet on ImageNet without it.
With $1$ attended location, top-$1$ accuracy drops from $73.12 \%$ (Table \ref{table_1}) to $65.21 \%$, while with $3$ and $5$ locations, accuracy drops from $74.41 \%$ and $74.62 \%$, to $67.33 \%$ and $68.55 \%$ respectively. The drop is substantial, placing TNet below both Saccader and BagNet-$77$ in terms of accuracy.

Per-feature regularization may have similar impact as cropping-based data augmentation, since it forces the model to make independent predictions with features from every attended location.
However, the attention policy is not random, but learned, which is crucial for the quality of the crops.
In addition, we don't get one image crop per epoch, but multiple crops in the same training iteration.
We hypothesize that this is important to prevent feature co-adaptation, since the model learns to recognize the same object from multiple crops simultaneously.

\section{Conclusion}
\label{sec6}
We proposed a novel multi-scale hard-attention architecture, TNet, that can efficiently scale to images of high resolution. By controlling the number of attended locations, TNet can adjust the accuracy-computation trade-off dynamically.
We demonstrated the efficacy of our method on ImageNet against strong hard-attention baselines, and we further verified its behavior with high-resolution satellite images (fMoW). The attention policy reveals the image regions deemed more informative by our model, and makes its predictions inherently interpretable. 

There are multiple research directions that can address current limitations of our method.
First, we would like the decision on the number of attended locations to stem from a content-dependent learned policy.
In addition, we would like scale-space traversal to be bi-directional, instead of merely top-down, in order for processing to be more adaptive.
To the same end, we would like already extracted features to condition the processing of subsequent locations.

On the broader impact of our approach, we hypothesize that under a number of assumptions, hard-attention has the potential to be useful to any intelligent agent that navigates through the immense complexity of the natural visual world.
These assumptions are that (1) the available resources are limited (2) there are performance constraints e.g. maximum response time, (3) the available information is practically infinite (4) all bits of information are not equally useful, and can even be misleading, e.g. noise.
In this context, it would be beneficial for an intelligent agent to prioritize the expenditure of its resources according to the utility of the available information, in order to  reach its performance goals; this is what a learned hard-attention mechanism can facilitate.

\bibliographystyle{plainnat}
\bibliography{neurips_2021}

\begin{thebibliography}{87}
\providecommand{\natexlab}[1]{#1}
\providecommand{\url}[1]{\texttt{#1}}
\expandafter\ifx\csname urlstyle\endcsname\relax
  \providecommand{\doi}[1]{doi: #1}\else
  \providecommand{\doi}{doi: \begingroup \urlstyle{rm}\Url}\fi

\bibitem[Ancona et~al.(2018)Ancona, Ceolini, {\"O}ztireli, and
  Gross]{ancona2017towards}
Marco Ancona, Enea Ceolini, Cengiz {\"O}ztireli, and Markus Gross.
\newblock Towards better understanding of gradient-based attribution methods
  for deep neural networks.
\newblock In \emph{International Conference on Learning Representations}, 2018.

\bibitem[Ba et~al.(2014)Ba, Mnih, and Kavukcuoglu]{ba2014multiple}
Jimmy Ba, Volodymyr Mnih, and Koray Kavukcuoglu.
\newblock Multiple object recognition with visual attention.
\newblock \emph{arXiv preprint arXiv:1412.7755}, 2014.

\bibitem[Ba et~al.(2016)Ba, Hinton, Mnih, Leibo, and Ionescu]{ba2016using}
Jimmy Ba, Geoffrey~E Hinton, Volodymyr Mnih, Joel~Z Leibo, and Catalin Ionescu.
\newblock Using fast weights to attend to the recent past.
\newblock In \emph{Advances in Neural Information Processing Systems}, pages
  4331--4339, 2016.

\bibitem[Bahdanau et~al.(2014)Bahdanau, Cho, and Bengio]{bahdanau2014neural}
Dzmitry Bahdanau, Kyunghyun Cho, and Yoshua Bengio.
\newblock Neural machine translation by jointly learning to align and
  translate.
\newblock \emph{arXiv preprint arXiv:1409.0473}, 2014.

\bibitem[Bello(2021)]{bello2021lambdanetworks}
Irwan Bello.
\newblock Lambdanetworks: Modeling long-range interactions without attention.
\newblock \emph{arXiv preprint arXiv:2102.08602}, 2021.

\bibitem[Bello et~al.(2019)Bello, Zoph, Vaswani, Shlens, and
  Le]{bello2019attention}
Irwan Bello, Barret Zoph, Ashish Vaswani, Jonathon Shlens, and Quoc~V Le.
\newblock Attention augmented convolutional networks.
\newblock In \emph{Proceedings of the IEEE/CVF International Conference on
  Computer Vision}, pages 3286--3295, 2019.

\bibitem[Brendel and Bethge(2019)]{brendel2019approximating}
Wieland Brendel and Matthias Bethge.
\newblock Approximating cnns with bag-of-local-features models works
  surprisingly well on imagenet.
\newblock \emph{arXiv preprint arXiv:1904.00760}, 2019.

\bibitem[Carion et~al.(2020)Carion, Massa, Synnaeve, Usunier, Kirillov, and
  Zagoruyko]{carion2020end}
Nicolas Carion, Francisco Massa, Gabriel Synnaeve, Nicolas Usunier, Alexander
  Kirillov, and Sergey Zagoruyko.
\newblock End-to-end object detection with transformers.
\newblock In \emph{European Conference on Computer Vision}, pages 213--229.
  Springer, 2020.

\bibitem[Chen et~al.(2017)Chen, Papandreou, Schroff, and
  Adam]{chen2017rethinking}
Liang-Chieh Chen, George Papandreou, Florian Schroff, and Hartwig Adam.
\newblock Rethinking atrous convolution for semantic image segmentation.
\newblock \emph{arXiv preprint arXiv:1706.05587}, 2017.

\bibitem[Chen et~al.(2018{\natexlab{a}})Chen, Papandreou, Kokkinos, Murphy, and
  Yuille]{chen2018deeplab}
Liang-Chieh Chen, George Papandreou, Iasonas Kokkinos, Kevin Murphy, and Alan~L
  Yuille.
\newblock Deeplab: Semantic image segmentation with deep convolutional nets,
  atrous convolution, and fully connected crfs.
\newblock \emph{IEEE transactions on pattern analysis and machine
  intelligence}, 40\penalty0 (4):\penalty0 834--848, 2018{\natexlab{a}}.

\bibitem[Chen et~al.(2018{\natexlab{b}})Chen, Kalantidis, Li, Yan, and
  Feng]{chen20182}
Yunpeng Chen, Yannis Kalantidis, Jianshu Li, Shuicheng Yan, and Jiashi Feng.
\newblock A$^{2}$-nets: Double attention networks.
\newblock \emph{arXiv preprint arXiv:1810.11579}, 2018{\natexlab{b}}.

\bibitem[Christie et~al.(2018)Christie, Fendley, Wilson, and
  Mukherjee]{fmow2018}
Gordon Christie, Neil Fendley, James Wilson, and Ryan Mukherjee.
\newblock Functional map of the world.
\newblock In \emph{CVPR}, 2018.

\bibitem[Cubuk et~al.(2020)Cubuk, Zoph, Shlens, and Le]{cubuk2020randaugment}
Ekin~D Cubuk, Barret Zoph, Jonathon Shlens, and Quoc~V Le.
\newblock Randaugment: Practical automated data augmentation with a reduced
  search space.
\newblock In \emph{Proceedings of the IEEE/CVF Conference on Computer Vision
  and Pattern Recognition Workshops}, pages 702--703, 2020.

\bibitem[Deng et~al.(2009)Deng, Dong, Socher, Li, Li, and
  Fei-Fei]{deng2009imagenet}
Jia Deng, Wei Dong, Richard Socher, Li-Jia Li, Kai Li, and Li~Fei-Fei.
\newblock Imagenet: A large-scale hierarchical image database.
\newblock In \emph{2009 IEEE conference on computer vision and pattern
  recognition}, pages 248--255. Ieee, 2009.

\bibitem[Denil et~al.(2012)Denil, Bazzani, Larochelle, and
  de~Freitas]{denil2012learning}
Misha Denil, Loris Bazzani, Hugo Larochelle, and Nando de~Freitas.
\newblock Learning where to attend with deep architectures for image tracking.
\newblock \emph{Neural computation}, 24\penalty0 (8):\penalty0 2151--2184,
  2012.

\bibitem[Devlin et~al.(2018)Devlin, Chang, Lee, and Toutanova]{devlin2018bert}
Jacob Devlin, Ming-Wei Chang, Kenton Lee, and Kristina Toutanova.
\newblock Bert: Pre-training of deep bidirectional transformers for language
  understanding.
\newblock \emph{arXiv preprint arXiv:1810.04805}, 2018.

\bibitem[Dosovitskiy et~al.(2020)Dosovitskiy, Beyer, Kolesnikov, Weissenborn,
  Zhai, Unterthiner, Dehghani, Minderer, Heigold, Gelly,
  et~al.]{dosovitskiy2020image}
Alexey Dosovitskiy, Lucas Beyer, Alexander Kolesnikov, Dirk Weissenborn,
  Xiaohua Zhai, Thomas Unterthiner, Mostafa Dehghani, Matthias Minderer, Georg
  Heigold, Sylvain Gelly, et~al.
\newblock An image is worth 16x16 words: Transformers for image recognition at
  scale.
\newblock \emph{arXiv preprint arXiv:2010.11929}, 2020.

\bibitem[Eigen et~al.(2014)Eigen, Puhrsch, and Fergus]{eigen2014depth}
David Eigen, Christian Puhrsch, and Rob Fergus.
\newblock Depth map prediction from a single image using a multi-scale deep
  network.
\newblock In \emph{Advances in neural information processing systems}, pages
  2366--2374, 2014.

\bibitem[Elsayed et~al.(2019)Elsayed, Kornblith, and Le]{elsayed2019saccader}
Gamaleldin~F Elsayed, Simon Kornblith, and Quoc~V Le.
\newblock Saccader: improving accuracy of hard attention models for vision.
\newblock \emph{arXiv preprint arXiv:1908.07644}, 2019.

\bibitem[Eslami et~al.(2016)Eslami, Heess, Weber, Tassa, Szepesvari, Hinton,
  et~al.]{eslami2016attend}
SM~Ali Eslami, Nicolas Heess, Theophane Weber, Yuval Tassa, David Szepesvari,
  Geoffrey~E Hinton, et~al.
\newblock Attend, infer, repeat: Fast scene understanding with generative
  models.
\newblock In \emph{Advances in Neural Information Processing Systems}, pages
  3225--3233, 2016.

\bibitem[Figurnov et~al.(2017)Figurnov, Collins, Zhu, Zhang, Huang, Vetrov, and
  Salakhutdinov]{figurnov2017spatially}
Michael Figurnov, Maxwell~D Collins, Yukun Zhu, Li~Zhang, Jonathan Huang,
  Dmitry Vetrov, and Ruslan Salakhutdinov.
\newblock Spatially adaptive computation time for residual networks.
\newblock In \emph{Proceedings of the IEEE Conference on Computer Vision and
  Pattern Recognition}, pages 1039--1048, 2017.

\bibitem[Glorot and Bengio(2010)]{glorot2010understanding}
Xavier Glorot and Yoshua Bengio.
\newblock Understanding the difficulty of training deep feedforward neural
  networks.
\newblock In \emph{Proceedings of the thirteenth international conference on
  artificial intelligence and statistics}, pages 249--256, 2010.

\bibitem[google research(2019)]{saccader2021github}
google research.
\newblock \emph{GitHub repository that contains the official Saccader and DRAM
  implementations}, 2019.
\newblock URL
  \url{https://github.com/google-research/google-research/tree/master/saccader}.

\bibitem[Gregor et~al.(2015)Gregor, Danihelka, Graves, Rezende, and
  Wierstra]{gregor2015draw}
Karol Gregor, Ivo Danihelka, Alex Graves, Danilo Rezende, and Daan Wierstra.
\newblock Draw: A recurrent neural network for image generation.
\newblock In \emph{International Conference on Machine Learning}, pages
  1462--1471, 2015.

\bibitem[He et~al.(2021)He, Chen, Liu, Kortylewski, Yang, Bai, Wang, and
  Yuille]{he2021transfg}
Ju~He, Jie-Neng Chen, Shuai Liu, Adam Kortylewski, Cheng Yang, Yutong Bai,
  Changhu Wang, and Alan Yuille.
\newblock Transfg: A transformer architecture for fine-grained recognition.
\newblock \emph{arXiv preprint arXiv:2103.07976}, 2021.

\bibitem[He et~al.(2014)He, Zhang, Ren, and Sun]{he2014spatial}
Kaiming He, Xiangyu Zhang, Shaoqing Ren, and Jian Sun.
\newblock Spatial pyramid pooling in deep convolutional networks for visual
  recognition.
\newblock In \emph{European conference on computer vision}, pages 346--361.
  Springer, 2014.

\bibitem[He et~al.(2016)He, Zhang, Ren, and Sun]{he2016deep}
Kaiming He, Xiangyu Zhang, Shaoqing Ren, and Jian Sun.
\newblock Deep residual learning for image recognition.
\newblock In \emph{Proceedings of the IEEE conference on computer vision and
  pattern recognition}, pages 770--778, 2016.

\bibitem[Hinton et~al.(2015)Hinton, Vinyals, and Dean]{hinton2015distilling}
Geoffrey Hinton, Oriol Vinyals, and Jeff Dean.
\newblock Distilling the knowledge in a neural network.
\newblock \emph{arXiv preprint arXiv:1503.02531}, 2015.

\bibitem[Howard(2013)]{howard2013some}
Andrew~G Howard.
\newblock Some improvements on deep convolutional neural network based image
  classification.
\newblock \emph{arXiv preprint arXiv:1312.5402}, 2013.

\bibitem[Howard et~al.(2017)Howard, Zhu, Chen, Kalenichenko, Wang, Weyand,
  Andreetto, and Adam]{howard2017mobilenets}
Andrew~G Howard, Menglong Zhu, Bo~Chen, Dmitry Kalenichenko, Weijun Wang,
  Tobias Weyand, Marco Andreetto, and Hartwig Adam.
\newblock Mobilenets: Efficient convolutional neural networks for mobile vision
  applications.
\newblock \emph{arXiv preprint arXiv:1704.04861}, 2017.

\bibitem[Hu et~al.(2018)Hu, Shen, and Sun]{hu2018squeeze}
Jie Hu, Li~Shen, and Gang Sun.
\newblock Squeeze-and-excitation networks.
\newblock In \emph{Proceedings of the IEEE conference on computer vision and
  pattern recognition}, pages 7132--7141, 2018.

\bibitem[Huang et~al.(2016)Huang, Sun, Liu, Sedra, and
  Weinberger]{huang2016deep}
Gao Huang, Yu~Sun, Zhuang Liu, Daniel Sedra, and Kilian~Q Weinberger.
\newblock Deep networks with stochastic depth.
\newblock In \emph{European conference on computer vision}, pages 646--661.
  Springer, 2016.

\bibitem[Huang et~al.(2017{\natexlab{a}})Huang, Chen, Li, Wu, van~der Maaten,
  and Weinberger]{huang2017multi}
Gao Huang, Danlu Chen, Tianhong Li, Felix Wu, Laurens van~der Maaten, and
  Kilian~Q Weinberger.
\newblock Multi-scale dense networks for resource efficient image
  classification.
\newblock \emph{arXiv preprint arXiv:1703.09844}, 2017{\natexlab{a}}.

\bibitem[Huang et~al.(2017{\natexlab{b}})Huang, Liu, Van Der~Maaten, and
  Weinberger]{huang2017densely}
Gao Huang, Zhuang Liu, Laurens Van Der~Maaten, and Kilian~Q Weinberger.
\newblock Densely connected convolutional networks.
\newblock In \emph{Proceedings of the IEEE conference on computer vision and
  pattern recognition}, pages 4700--4708, 2017{\natexlab{b}}.

\bibitem[Iandola et~al.(2016)Iandola, Han, Moskewicz, Ashraf, Dally, and
  Keutzer]{iandola2016squeezenet}
Forrest~N Iandola, Song Han, Matthew~W Moskewicz, Khalid Ashraf, William~J
  Dally, and Kurt Keutzer.
\newblock Squeezenet: Alexnet-level accuracy with 50x fewer parameters and< 0.5
  mb model size.
\newblock \emph{arXiv preprint arXiv:1602.07360}, 2016.

\bibitem[Ioffe and Szegedy(2015)]{ioffe2015batch}
Sergey Ioffe and Christian Szegedy.
\newblock Batch normalization: Accelerating deep network training by reducing
  internal covariate shift.
\newblock In \emph{International conference on machine learning}, pages
  448--456. PMLR, 2015.

\bibitem[Itti et~al.(1998)Itti, Koch, and Niebur]{itti1998model}
Laurent Itti, Christof Koch, and Ernst Niebur.
\newblock A model of saliency-based visual attention for rapid scene analysis.
\newblock \emph{IEEE Transactions on pattern analysis and machine
  intelligence}, 20\penalty0 (11):\penalty0 1254--1259, 1998.

\bibitem[Jaderberg et~al.(2014)Jaderberg, Vedaldi, and
  Zisserman]{jaderberg2014speeding}
Max Jaderberg, Andrea Vedaldi, and Andrew Zisserman.
\newblock Speeding up convolutional neural networks with low rank expansions.
\newblock \emph{arXiv preprint arXiv:1405.3866}, 2014.

\bibitem[Katharopoulos and Fleuret(2019)]{katharopoulos2019processing}
Angelos Katharopoulos and Fran{\c{c}}ois Fleuret.
\newblock Processing megapixel images with deep attention-sampling models.
\newblock \emph{arXiv preprint arXiv:1905.03711}, 2019.

\bibitem[Ke et~al.(2017)Ke, Maire, and Yu]{ke2017multigrid}
Tsung-Wei Ke, Michael Maire, and Stella~X Yu.
\newblock Multigrid neural architectures.
\newblock In \emph{Proceedings of the IEEE Conference on Computer Vision and
  Pattern Recognition}, pages 6665--6673, 2017.

\bibitem[Kingma and Ba(2014)]{kingma2014adam}
Diederik~P Kingma and Jimmy Ba.
\newblock Adam: A method for stochastic optimization.
\newblock \emph{arXiv preprint arXiv:1412.6980}, 2014.

\bibitem[Larochelle and Hinton(2010)]{larochelle2010learning}
Hugo Larochelle and Geoffrey~E Hinton.
\newblock Learning to combine foveal glimpses with a third-order boltzmann
  machine.
\newblock In \emph{Advances in neural information processing systems}, pages
  1243--1251, 2010.

\bibitem[Larsson et~al.(2016)Larsson, Maire, and
  Shakhnarovich]{larsson2016fractalnet}
Gustav Larsson, Michael Maire, and Gregory Shakhnarovich.
\newblock Fractalnet: Ultra-deep neural networks without residuals.
\newblock \emph{arXiv preprint arXiv:1605.07648}, 2016.

\bibitem[LeCun et~al.(1990)LeCun, Denker, and Solla]{lecun1990optimal}
Yann LeCun, John~S Denker, and Sara~A Solla.
\newblock Optimal brain damage.
\newblock In \emph{Advances in neural information processing systems}, pages
  598--605, 1990.

\bibitem[Levi and Ullman(2018)]{levi2018efficient}
Hila Levi and Shimon Ullman.
\newblock Efficient coarse-to-fine non-local module for the detection of small
  objects.
\newblock \emph{arXiv preprint arXiv:1811.12152}, 2018.

\bibitem[Lin et~al.(2017)Lin, Doll{\'a}r, Girshick, He, Hariharan, and
  Belongie]{lin2017feature}
Tsung-Yi Lin, Piotr Doll{\'a}r, Ross~B Girshick, Kaiming He, Bharath Hariharan,
  and Serge~J Belongie.
\newblock Feature pyramid networks for object detection.
\newblock In \emph{CVPR}, volume~1, page~4, 2017.

\bibitem[Liu et~al.(2016)Liu, Anguelov, Erhan, Szegedy, Reed, Fu, and
  Berg]{liu2016ssd}
Wei Liu, Dragomir Anguelov, Dumitru Erhan, Christian Szegedy, Scott Reed,
  Cheng-Yang Fu, and Alexander~C Berg.
\newblock Ssd: Single shot multibox detector.
\newblock In \emph{European conference on computer vision}, pages 21--37.
  Springer, 2016.

\bibitem[Luo et~al.(2017)Luo, Li, Urtasun, and Zemel]{luo2017understanding}
Wenjie Luo, Yujia Li, Raquel Urtasun, and Richard Zemel.
\newblock Understanding the effective receptive field in deep convolutional
  neural networks.
\newblock \emph{arXiv preprint arXiv:1701.04128}, 2017.

\bibitem[Marra et~al.(2020)Marra, Gragnaniello, Verdoliva, and
  Poggi]{marra2020full}
Francesco Marra, Diego Gragnaniello, Luisa Verdoliva, and Giovanni Poggi.
\newblock A full-image full-resolution end-to-end-trainable cnn framework for
  image forgery detection.
\newblock \emph{IEEE Access}, 8:\penalty0 133488--133502, 2020.

\bibitem[Mnih et~al.(2014)Mnih, Heess, Graves, et~al.]{mnih2014recurrent}
Volodymyr Mnih, Nicolas Heess, Alex Graves, et~al.
\newblock Recurrent models of visual attention.
\newblock In \emph{Advances in neural information processing systems}, pages
  2204--2212, 2014.

\bibitem[Najibi et~al.(2018)Najibi, Singh, and Davis]{najibi2018autofocus}
Mahyar Najibi, Bharat Singh, and Larry~S Davis.
\newblock Autofocus: Efficient multi-scale inference.
\newblock \emph{arXiv preprint arXiv:1812.01600}, 2018.

\bibitem[Parmar et~al.(2018)Parmar, Vaswani, Uszkoreit, Kaiser, Shazeer, Ku,
  and Tran]{parmar2018image}
Niki Parmar, Ashish Vaswani, Jakob Uszkoreit, Lukasz Kaiser, Noam Shazeer,
  Alexander Ku, and Dustin Tran.
\newblock Image transformer.
\newblock In \emph{International Conference on Machine Learning}, pages
  4055--4064. PMLR, 2018.

\bibitem[Pinckaers et~al.(2019)Pinckaers, van Ginneken, and
  Litjens]{pinckaers2019streaming}
Hans Pinckaers, Bram van Ginneken, and Geert Litjens.
\newblock Streaming convolutional neural networks for end-to-end learning with
  multi-megapixel images.
\newblock \emph{arXiv preprint arXiv:1911.04432}, 2019.

\bibitem[Pinheiro and Collobert(2014)]{pinheiro2014recurrent}
Pedro~HO Pinheiro and Ronan Collobert.
\newblock Recurrent convolutional neural networks for scene labeling.
\newblock In \emph{31st International Conference on Machine Learning (ICML)},
  number EPFL-CONF-199822, 2014.

\bibitem[Ramachandran et~al.(2019)Ramachandran, Parmar, Vaswani, Bello,
  Levskaya, and Shlens]{ramachandran2019stand}
Prajit Ramachandran, Niki Parmar, Ashish Vaswani, Irwan Bello, Anselm Levskaya,
  and Jonathon Shlens.
\newblock Stand-alone self-attention in vision models.
\newblock \emph{arXiv preprint arXiv:1906.05909}, 2019.

\bibitem[Ramapuram et~al.(2018)Ramapuram, Diephuis, Webb, and
  Kalousis]{ramapuram2018variational}
Jason Ramapuram, Maurits Diephuis, Russ Webb, and Alexandros Kalousis.
\newblock Variational saccading: Efficient inference for large resolution
  images.
\newblock \emph{arXiv preprint arXiv:1812.03170}, 2018.

\bibitem[Ranzato(2014)]{ranzato2014learning}
Marc'Aurelio Ranzato.
\newblock On learning where to look.
\newblock \emph{arXiv preprint arXiv:1405.5488}, 2014.

\bibitem[Rastegari et~al.(2016)Rastegari, Ordonez, Redmon, and
  Farhadi]{rastegari2016xnor}
Mohammad Rastegari, Vicente Ordonez, Joseph Redmon, and Ali Farhadi.
\newblock Xnor-net: Imagenet classification using binary convolutional neural
  networks.
\newblock In \emph{European Conference on Computer Vision}, pages 525--542.
  Springer, 2016.

\bibitem[Ronneberger et~al.(2015)Ronneberger, Fischer, and
  Brox]{ronneberger2015u}
Olaf Ronneberger, Philipp Fischer, and Thomas Brox.
\newblock U-net: Convolutional networks for biomedical image segmentation.
\newblock In \emph{International Conference on Medical image computing and
  computer-assisted intervention}, pages 234--241. Springer, 2015.

\bibitem[Selvaraju et~al.(2017)Selvaraju, Cogswell, Das, Vedantam, Parikh, and
  Batra]{selvaraju2017grad}
Ramprasaath~R Selvaraju, Michael Cogswell, Abhishek Das, Ramakrishna Vedantam,
  Devi Parikh, and Dhruv Batra.
\newblock Grad-cam: Visual explanations from deep networks via gradient-based
  localization.
\newblock In \emph{Proceedings of the IEEE international conference on computer
  vision}, pages 618--626, 2017.

\bibitem[Sermanet et~al.(2014)Sermanet, Frome, and Real]{sermanet2014attention}
Pierre Sermanet, Andrea Frome, and Esteban Real.
\newblock Attention for fine-grained categorization.
\newblock \emph{arXiv preprint arXiv:1412.7054}, 2014.

\bibitem[Shen et~al.(2020)Shen, Wu, Phang, Park, Liu, Tyagi, Heacock, Kim, Moy,
  Cho, et~al.]{shen2020interpretable}
Yiqiu Shen, Nan Wu, Jason Phang, Jungkyu Park, Kangning Liu, Sudarshini Tyagi,
  Laura Heacock, S~Gene Kim, Linda Moy, Kyunghyun Cho, et~al.
\newblock An interpretable classifier for high-resolution breast cancer
  screening images utilizing weakly supervised localization.
\newblock \emph{Medical Image Analysis}, page 101908, 2020.

\bibitem[Shen et~al.(2021)Shen, Wu, Phang, Park, Liu, Tyagi, Heacock, Kim, Moy,
  Cho, et~al.]{shen2021interpretable}
Yiqiu Shen, Nan Wu, Jason Phang, Jungkyu Park, Kangning Liu, Sudarshini Tyagi,
  Laura Heacock, S~Gene Kim, Linda Moy, Kyunghyun Cho, et~al.
\newblock An interpretable classifier for high-resolution breast cancer
  screening images utilizing weakly supervised localization.
\newblock \emph{Medical image analysis}, 68:\penalty0 101908, 2021.

\bibitem[Srivastava et~al.(2014)Srivastava, Hinton, Krizhevsky, Sutskever, and
  Salakhutdinov]{srivastava2014dropout}
Nitish Srivastava, Geoffrey Hinton, Alex Krizhevsky, Ilya Sutskever, and Ruslan
  Salakhutdinov.
\newblock Dropout: a simple way to prevent neural networks from overfitting.
\newblock \emph{The journal of machine learning research}, 15\penalty0
  (1):\penalty0 1929--1958, 2014.

\bibitem[Szegedy et~al.(2015)Szegedy, Liu, Jia, Sermanet, Reed, Anguelov,
  Erhan, Vanhoucke, and Rabinovich]{szegedy2015going}
Christian Szegedy, Wei Liu, Yangqing Jia, Pierre Sermanet, Scott Reed, Dragomir
  Anguelov, Dumitru Erhan, Vincent Vanhoucke, and Andrew Rabinovich.
\newblock Going deeper with convolutions.
\newblock In \emph{Proceedings of the IEEE conference on computer vision and
  pattern recognition}, pages 1--9, 2015.

\bibitem[Tan and Le(2019)]{tan2019efficientnet}
Mingxing Tan and Quoc Le.
\newblock Efficientnet: Rethinking model scaling for convolutional neural
  networks.
\newblock In \emph{International Conference on Machine Learning}, pages
  6105--6114. PMLR, 2019.

\bibitem[Tan and Le(2021)]{tan2021efficientnetv2}
Mingxing Tan and Quoc~V Le.
\newblock Efficientnetv2: Smaller models and faster training.
\newblock \emph{arXiv preprint arXiv:2104.00298}, 2021.

\bibitem[Tan et~al.(2019)Tan, Chen, Pang, Vasudevan, Sandler, Howard, and
  Le]{tan2019mnasnet}
Mingxing Tan, Bo~Chen, Ruoming Pang, Vijay Vasudevan, Mark Sandler, Andrew
  Howard, and Quoc~V Le.
\newblock Mnasnet: Platform-aware neural architecture search for mobile.
\newblock In \emph{Proceedings of the IEEE/CVF Conference on Computer Vision
  and Pattern Recognition}, pages 2820--2828, 2019.

\bibitem[tensorflow(2021)]{efficientnet2021github}
tensorflow.
\newblock \emph{GitHub repository that contains the official EfficientnNet
  implementation}, 2021.
\newblock URL
  \url{https://github.com/tensorflow/tpu/tree/master/models/official/efficientnet}.

\bibitem[Tompson et~al.(2015)Tompson, Goroshin, Jain, LeCun, and
  Bregler]{tompson2015efficient}
Jonathan Tompson, Ross Goroshin, Arjun Jain, Yann LeCun, and Christoph Bregler.
\newblock Efficient object localization using convolutional networks.
\newblock In \emph{Proceedings of the IEEE conference on computer vision and
  pattern recognition}, pages 648--656, 2015.

\bibitem[Uzkent and Ermon(2020)]{uzkent2020learning}
Burak Uzkent and Stefano Ermon.
\newblock Learning when and where to zoom with deep reinforcement learning.
\newblock In \emph{Proceedings of the IEEE/CVF Conference on Computer Vision
  and Pattern Recognition}, pages 12345--12354, 2020.

\bibitem[Uzkent et~al.(2019)Uzkent, Sheehan, Meng, Tang, Burke, Lobell, and
  Ermon]{uzkent2019learning}
Burak Uzkent, Evan Sheehan, Chenlin Meng, Zhongyi Tang, Marshall Burke, David
  Lobell, and Stefano Ermon.
\newblock Learning to interpret satellite images in global scale using
  wikipedia.
\newblock \emph{arXiv preprint arXiv:1905.02506}, 2019.

\bibitem[Van~Horn et~al.(2015)Van~Horn, Branson, Farrell, Haber, Barry,
  Ipeirotis, Perona, and Belongie]{van2015building}
Grant Van~Horn, Steve Branson, Ryan Farrell, Scott Haber, Jessie Barry, Panos
  Ipeirotis, Pietro Perona, and Serge Belongie.
\newblock Building a bird recognition app and large scale dataset with citizen
  scientists: The fine print in fine-grained dataset collection.
\newblock In \emph{Proceedings of the IEEE Conference on Computer Vision and
  Pattern Recognition}, pages 595--604, 2015.

\bibitem[Vaswani et~al.(2017)Vaswani, Shazeer, Parmar, Uszkoreit, Jones, Gomez,
  Kaiser, and Polosukhin]{vaswani2017attention}
Ashish Vaswani, Noam Shazeer, Niki Parmar, Jakob Uszkoreit, Llion Jones,
  Aidan~N Gomez, Lukasz Kaiser, and Illia Polosukhin.
\newblock Attention is all you need.
\newblock \emph{arXiv preprint arXiv:1706.03762}, 2017.

\bibitem[Wah et~al.(2011)Wah, Branson, Welinder, Perona, and
  Belongie]{wah2011caltech}
Catherine Wah, Steve Branson, Peter Welinder, Pietro Perona, and Serge
  Belongie.
\newblock The caltech-ucsd birds-200-2011 dataset.
\newblock 2011.

\bibitem[Wang et~al.(2019)Wang, Kembhavi, Farhadi, Yuille, and
  Rastegari]{wang2019elastic}
Huiyu Wang, Aniruddha Kembhavi, Ali Farhadi, Alan~L Yuille, and Mohammad
  Rastegari.
\newblock Elastic: Improving cnns with dynamic scaling policies.
\newblock In \emph{Proceedings of the IEEE Conference on Computer Vision and
  Pattern Recognition}, pages 2258--2267, 2019.

\bibitem[Wang et~al.(2017)Wang, Liu, and Foroosh]{wang2017factorized}
Min Wang, Baoyuan Liu, and Hassan Foroosh.
\newblock Factorized convolutional neural networks.
\newblock In \emph{Proceedings of the IEEE International Conference on Computer
  Vision}, pages 545--553, 2017.

\bibitem[Wang et~al.(2018)Wang, Girshick, Gupta, and He]{wang2018non}
Xiaolong Wang, Ross Girshick, Abhinav Gupta, and Kaiming He.
\newblock Non-local neural networks.
\newblock In \emph{Proceedings of the IEEE conference on computer vision and
  pattern recognition}, pages 7794--7803, 2018.

\bibitem[Williams(1992)]{williams1992simple}
Ronald~J Williams.
\newblock Simple statistical gradient-following algorithms for connectionist
  reinforcement learning.
\newblock \emph{Machine learning}, 8\penalty0 (3-4):\penalty0 229--256, 1992.

\bibitem[Wu et~al.(2018)Wu, Nagarajan, Kumar, Rennie, Davis, Grauman, and
  Feris]{wu2018blockdrop}
Zuxuan Wu, Tushar Nagarajan, Abhishek Kumar, Steven Rennie, Larry~S Davis,
  Kristen Grauman, and Rogerio Feris.
\newblock Blockdrop: Dynamic inference paths in residual networks.
\newblock In \emph{Proceedings of the IEEE Conference on Computer Vision and
  Pattern Recognition}, pages 8817--8826, 2018.

\bibitem[Xie et~al.(2020)Xie, Luong, Hovy, and Le]{xie2020self}
Qizhe Xie, Minh-Thang Luong, Eduard Hovy, and Quoc~V Le.
\newblock Self-training with noisy student improves imagenet classification.
\newblock In \emph{Proceedings of the IEEE/CVF Conference on Computer Vision
  and Pattern Recognition}, pages 10687--10698, 2020.

\bibitem[Xu et~al.(2015)Xu, Ba, Kiros, Cho, Courville, Salakhudinov, Zemel, and
  Bengio]{xu2015show}
Kelvin Xu, Jimmy Ba, Ryan Kiros, Kyunghyun Cho, Aaron Courville, Ruslan
  Salakhudinov, Rich Zemel, and Yoshua Bengio.
\newblock Show, attend and tell: Neural image caption generation with visual
  attention.
\newblock In \emph{International conference on machine learning}, pages
  2048--2057, 2015.

\bibitem[Yu and Koltun(2015)]{yu2015multi}
Fisher Yu and Vladlen Koltun.
\newblock Multi-scale context aggregation by dilated convolutions.
\newblock \emph{arXiv preprint arXiv:1511.07122}, 2015.

\bibitem[Yu et~al.(2018)Yu, Li, Chen, Lai, Morariu, Han, Gao, Lin, and
  Davis]{yu2018nisp}
Ruichi Yu, Ang Li, Chun-Fu Chen, Jui-Hsin Lai, Vlad~I Morariu, Xintong Han,
  Mingfei Gao, Ching-Yung Lin, and Larry~S Davis.
\newblock Nisp: Pruning networks using neuron importance score propagation.
\newblock In \emph{Proceedings of the IEEE Conference on Computer Vision and
  Pattern Recognition}, pages 9194--9203, 2018.

\bibitem[Zamir et~al.(2017)Zamir, Wu, Sun, Shen, Shi, Malik, and
  Savarese]{zamir2017feedback}
Amir~R Zamir, Te-Lin Wu, Lin Sun, William~B Shen, Bertram~E Shi, Jitendra
  Malik, and Silvio Savarese.
\newblock Feedback networks.
\newblock In \emph{Proceedings of the IEEE Conference on Computer Vision and
  Pattern Recognition}, pages 1308--1317, 2017.

\bibitem[Zhao et~al.(2020)Zhao, Jia, and Koltun]{zhao2020exploring}
Hengshuang Zhao, Jiaya Jia, and Vladlen Koltun.
\newblock Exploring self-attention for image recognition.
\newblock In \emph{Proceedings of the IEEE/CVF Conference on Computer Vision
  and Pattern Recognition}, pages 10076--10085, 2020.

\bibitem[Zhuang et~al.(2020)Zhuang, Wang, and Qiao]{zhuang2020learning}
Peiqin Zhuang, Yali Wang, and Yu~Qiao.
\newblock Learning attentive pairwise interaction for fine-grained
  classification.
\newblock In \emph{Proceedings of the AAAI Conference on Artificial
  Intelligence}, volume~34, pages 13130--13137, 2020.

\end{thebibliography}

\newpage

\title{Appendix}
\secondTitle

\renewcommand\thesection{\Alph{section}}
\renewcommand{\theHsection}{appendixsection.\Alph{section}}
\setcounter{section}{0}

\section{Training}
\label{App_sec1}

\subsection{Learning rule derivation}
\label{App_sec1_1}
The \emph{REINFORCE} rule naturally emerges if we optimize the log likelihood of the labels, while considering the attended locations as latent variables \cite{ba2014multiple}.
Given a batch of $N$ images, for the log likelihood we get:
\begin{equation}
\label{App_eq1}
\sum_{i=1}^{N} \log{p(y_i|x_i, w)} = \sum_{i=1}^{N} \log{\sum_{l^i}{p(l^i|x_i, w) p(y_i|l^i, x_i, w)}}
\end{equation}

where $x_i$ is the $i$-th image in the batch, $y_i$ is its label, and $w$ are the parameters of our model.
$p(l^i|x_i, w)$ is the probability that the sequence of locations $l^i$ is attended for image $x_i$, and $p(y_i|l^i, x_i, w)$ is the probability of predicting the correct label after attending to $l^i$.
Equation \ref{App_eq1} describes the log likelihood of the labels in terms of all location sequences that could be attended.
$p(y_i|l^i, x_i, w)$ is computed by the classification module, and $p(l^{i}|x_i, w)$ is computed by the location module (see Section \ref{App_sec1_2}).

We use Jensen's inequality in Equation \ref{App_eq1} to derive the following lower bound on the log likelihood:
\begin{equation}
\label{App_eq2}
\sum_{i=1}^{N} \log{p(y_i|x_i, w)} \geq \sum_{i=1}^{N} \sum_{l^i}{p(l^i|x_i, w) \log{p(y_i|l^i, x_i, w)}} = F
\end{equation}

By maximizing the lower bound $F$, we expect to maximize the log likelihood. The update rule we use, is the partial derivative of $F$ with respect to $w$, normalized by the number of images in the batch. We get:
\begin{align}
\frac{1}{N} \frac{\partial F}{\partial w} &= \frac{1}{N} \sum_{i=1}^{N} \sum_{l^i} \Big[ p(l^i|x_i, w) \frac{\partial \log{p(y_i|l^i, x_i, w)}}{\partial w} + \log{p(y_i|l^i, x_i, w)} \frac{\partial p(l^i|x_i, w)}{\partial w} \Big] \Rightarrow \nonumber \\
\label{App_eq3}
\frac{1}{N} \frac{\partial F}{\partial w} &= \frac{1}{N} \sum_{i=1}^{N} \sum_{l^i} p(l^i|x_i, w) \Big[ \frac{\partial \log{p(y_i|l^i, x_i, w)}}{\partial w} + \log{p(y_i|l^i, x_i, w)} \frac{\partial \log{p(l^i|x_i, w)}}{\partial w} \Big]
\end{align}

To derive \eqref{App_eq3}, we used the log derivative trick. As we can see, for each image $x_i$ we need to calculate an expectation according to $p(l^i|x_i, w)$. We approximate each expectation with a Monte Carlo estimator of $M$ samples:
\begin{align}
\frac{1}{N} \frac{\partial F}{\partial w} \approx \frac{1}{N} \frac{\partial \tilde{F}}{\partial w} = \frac{1}{N} \sum_{i=1}^{N} \frac{1}{M} \sum_{m=1}^{M}& \Big[ \frac{\partial \log{p(y_i|l^{i,m}, x_i, w)}}{\partial w} + \nonumber \\
\label{App_eq4}
&\log{p(y_i|l^{i,m}, x_i, w)} \frac{\partial \log{p(l^{i,m}|x_i, w)}}{\partial w} \Big]
\end{align}

$l^{i,m}$ is the sequence of locations attended during the $m$-th sample from $p(l^i|x_i, w)$ (we get samples by repeating the processing of image $x_i$).

In order to reduce the variance of the estimator, we replace $\log{p(y_i|l^{i,m}, x_i, w)}$ with a reward function $R_{i,m}$, which is equal to $1$ when the prediction for $x_i$ in the $m$-th sample is correct, and $0$ otherwise.
In addition, we use the baseline technique from \cite{xu2015show}, which corresponds to the exponential moving average of the mean reward $R_{i,m} \; \forall i, m$, and is updated after processing each training batch. Our baseline is initialized to $0.5$, and after the $n$-th batch we get:
\begin{equation}
\label{App_eq5}
b_n = 0.9 \cdot b_{n-1} + 0.1 \cdot \frac{1}{NM} \sum_{i=1}^{NM} R_i^n
\end{equation}

where $R_i^n$ is the reward for the $i$-th image in the $n$-th batch. Since we use $M$ samples for the Monte Carlo estimator of each image, we simply consider that our batch has size $N M$ to simplify notation.
Our learning rule \eqref{App_eq4} is updated as follows:
\begin{equation}
\label{App_eq6}
L_F = \frac{1}{NM} \sum_{i=1}^{NM} \Big[ \frac{\partial \log{p(y_i|l^{i}, x_i, w)}}{\partial w} + \lambda_f (R_i - b) \frac{\partial \log{p(l^{i}|x_i, w)}}{\partial w} \Big]
\end{equation}

For simplicity, we drop the subscript of $b_n$ that indicates the batch we are processing.
Also, we add a weighting hyperparameter $\lambda_f$.
Equation \ref{App_eq6} is the learning rule we present in Section \ref{sec4_1}, and this concludes our derivation.

\subsection{Sampling approximation}
\label{App_sec1_2}
In order to attend to a sequence of locations $l^i$, we sample without replacement from a series of Categorical distributions. For the probability of attending to a sequence $l^i$, we get:
\begin{equation}
\label{App_eq7}
p(l^i|x_i, w) = \prod_{j=1}^{N^{l^i}}{\prod_{r=1}^{L^{l^i}_{j}}{\prod_{k=1}^{g}{ \Big[ \frac{p^{l^i}_{j}(l_{k}|x_i, w)^{u_{j,k,r}^{l^i}}}{\sum_{k=1}^{g}{ \big[ p^{l^i}_{j}(l_{k}|x_i, w) \prod_{r^{'}=1}^{r-1}{(1-u^{l^i}_{j,k,r^{'}})} \big]}} \Big] }}}
\end{equation}

where $N^{l^i}$ is the number of Categorical distributions (equal to the number of times the location module is applied), $L^{l^i}_j$ is the number of samples we draw from the $j$-th distribution, and $g$ is the total number of candidate locations per distribution.
In the example of Fig. \ref{fig_4}, we consider $3$ distributions ($N^{l^i} = 3$), $L^{l^i}_1=2$ in the $2$nd processing level and $L^{l^i}_2=L^{l^i}_3=1$ in the $3$rd, and $g=4$ since we consider a $2 \times 2$ grid.

$l_{k}$ is the $k$-th out of the $g$ candidate locations, and $p^{l^i}_{j}(l_{k}|x_i, w)$ is the probability of selecting $l_{k}$ in the $j$-th distribution.
$u_{j,k,r}^{l^i}$ is an indicator function that is equal to $1$ when location $l_{k}$ is attended as the \mbox{$r$-th} sample of the $j$-th distribution, and $0$ otherwise.
$p^{l^i}_{j}(l_{k}|x_i, w)$ is computed by the location module, and $u_{j,k,r}^{l^i}$ is the outcome of sampling from the $j$-th Categorical distribution.
The denominator in \eqref{App_eq7} is applicable for $r > 1$, and normalizes the probabilities of the $j$-th Categorical distribution before the $r$-th sample, to account for the lack of replacement.

In order to simplify our implementation of sampling dictated by \eqref{App_eq7}, we introduce two modifications.
First, we approximate sampling by selecting the locations with the $L^{l^i}_j$ highest probabilities.
Potential downside is that we miss the opportunity to attend to less probable locations that may have valuable information (less exploration). However, at the beginning of training, all locations start with practically equal probability, and even by picking the top $L^{l^i}_j$ locations, we are able to explore the location space.
Second, we disregard the normalization factor for each $p^{l^i}_{j}(l_{k}|x_i, w)^{u_{j,k,r}^{l^i}}$ (denominator in \eqref{App_eq7}). This simplification does not affect the relative ordering between the probabilities of each Categorical distribution. As a result, the $L^{l^i}_j$ locations with the highest probabilities that we attend to, remain the same.

\section{Experimental evaluation}
\label{App_sec2}

\subsection{Architectures}
\label{App_sec_2_1}
\setlength{\tabcolsep}{8pt}
\begin{table*}[!t]
\caption{Building blocks of our architectures. ConvBlock and MBConv$F$ have residual connections that add the input to the output. If $s > 1$, or the number of input channels is not equal to the output channels, MBConv$F$ drops the residual connection.
If the same conditions hold true for the ConvBlock, it applies an $1 \times 1$ convolution with stride $s$ and channels $C$ to the input before it is added to the output. Also, if $p$ is VALID, a total margin of $k-1$ pixels is dropped from each spatial dimension of the input, before it is passed through the residual connection.
The first layer in MBConv$F$ is performed only if $F \neq 1$. Both for SE-($C_r, \; r$) and MBConv$F$, $C_{in}$ corresponds to the number of input channels and is not a parameter of the blocks. Batch Norm \cite{ioffe2015batch} is applied before the activation.
GAP stands for Global Average Pooling, and DWConv for depthwise convolution.}
\label{App_table_1}
\begin{center}
\resizebox{0.9999\textwidth}{!}{
\begin{small}
\begin{tabular}{cccccccc}
\toprule
\multirow{2}{*}{\textbf{Block Type}} & \multirow{2}{*}{\shortstack{\textbf{Layer/Block} \\ \textbf{Type}}} & \multirow{2}{*}{\shortstack{\textbf{Kernel} \\ \textbf{Size}}} & \multirow{2}{*}{\shortstack{\textbf{\#Output} \\ \textbf{Channels}}} & \multirow{2}{*}{\textbf{Stride}} & \multirow{2}{*}{\textbf{Padding}} & \multirow{2}{*}{\shortstack{\textbf{Batch} \\ \textbf{Norm}}} & \multirow{2}{*}{\textbf{Activation}} \\
& & & & & & & \\
\midrule
\multirow{3}{*}{\textbf{ConvBlock} \cite{he2016deep}} & Conv & $1 \times 1$ & $C / 4$ & $1$ & SAME & - & Leaky ReLU \\
& Conv & $k \times k$ & $C / 4$ & $s$ & $p$ & - & Leaky ReLU \\
& Conv & $1 \times 1$ & $C$ & $1$ & SAME & - & Leaky ReLU \\
\midrule
\multirow{4}{*}{\shortstack{\textbf{Squeeze and} \\ \textbf{Excitation-($C_r, \; r$)} \\ \textbf{SE-($C_r, \; r$)} \cite{hu2018squeeze}}} & GAP & - & $C_{in}$ & - & - & - & - \\
& Conv & $1 \times 1$ & $C_r \cdot r$ & $1$ & SAME & - & SiLU \\
& Conv & $1 \times 1$ & $C_{in}$ & $1$ & SAME & - & Sigmoid \\
& Multiply & - & $C_{in}$ & - & - & - & - \\
\midrule
\multirow{5}{*}{\textbf{MBConv$F$} \cite{tan2019mnasnet}} & \multirow{2}{*}{\shortstack{Conv \\ (if $F \neq 1$)}} & \multirow{2}{*}{$1 \times 1$} & \multirow{2}{*}{$C_{in} \cdot F$} & \multirow{2}{*}{$1$} & \multirow{2}{*}{SAME} & \multirow{2}{*}{$\surd$} & \multirow{2}{*}{SiLU} \\
& & & & & & \\
\cmidrule{2-8}
& DWConv & $k \times k$ & $C_{in} \cdot F$ & $s$ & SAME & $\surd$ & SiLU \\
& SE-($C_{in}, \; 0.25$) & - & $C_{in} \cdot F$ & - & - & - & - \\
& Conv & $1 \times 1$ & $C$ & $1$ & SAME & $\surd$ & - \\
\bottomrule
\end{tabular}
\end{small}
}
\end{center}
\vskip -0.1in
\end{table*}
\setlength{\tabcolsep}{8pt}
\begin{table*}[!t]
\caption{TNet architecture used on ImageNet (see Section \ref{sec5_1}).
BagNet-$77$ baseline corresponds to the feature extraction module followed by the classification module.
For simplicity, we provide only the spatial dimensions (without the channel dimension) of the feature extraction module's output.
The location module receives two inputs and combines them into a single input feature map of size $5 \times 5 \times 1538$ (see Section~\ref{App_sec_2_1_1}).
The positional encoding module receives a feature vector and a positional encoding vector, and concatenates them to an $1 \times 1024$ input vector (see Section~\ref{App_sec_2_1_2}).}
\label{App_table_2}
\begin{center}
\resizebox{1.0\textwidth}{!}{
\begin{small}
\begin{tabular}{ccccccccc}
\toprule
\multirow{2}{*}{\textbf{Module}} & \textbf{Layer/Block} & \textbf{Kernel} & \multirow{2}{*}{\shortstack{\textbf{\#Output} \\ \textbf{Channels}}} & \multirow{2}{*}{\textbf{Stride}} & \multirow{2}{*}{\textbf{Padding}} & \multirow{2}{*}{\textbf{Activation}} & \textbf{Output} & \textbf{Receptive} \\
& \textbf{Type} & \textbf{Size} & & & & & \textbf{Size} & \textbf{Field} \\
\midrule
\multirow{18}{*}{\shortstack{\textbf{Feature} \\ \textbf{Extraction}}} & Input & - & - & - & - & - & $77 \times 77$ & - \\
\cmidrule{2-9}
& Conv & $3 \times 3$ & $64$ & $1$ & VALID & Leaky ReLU & $75 \times 75$ & $3 \times 3$ \\
\cmidrule{2-9}
& ConvBlock & $3 \times 3$ & $256$ & $2$ & SAME & - & $38 \times 38$ & $5 \times 5$ \\
& ConvBlock & $3 \times 3$ & $256$ & $1$ & SAME & - & $38 \times 38$ & $9 \times 9$ \\
& ConvBlock & $1 \times 1$ & $256$ & $1$ & SAME & - & $38 \times 38$ & $9 \times 9$ \\
\cmidrule{2-9}
& ConvBlock & $3 \times 3$ & $512$ & $2$ & SAME & - & $19 \times 19$ & $13 \times 13$ \\
& ConvBlock & $3 \times 3$ & $512$ & $1$ & SAME & - & $19 \times 19$ & $21 \times 21$ \\
& ConvBlock ($\times 2$) & $1 \times 1$ & $512$ & $1$ & SAME & - & $19 \times 19$ & $21 \times 21$ \\
\cmidrule{2-9}
& ConvBlock & $3 \times 3$ & $1024$ & $2$ & VALID & - & $9 \times 9$ & $29 \times 29$ \\
& ConvBlock & $3 \times 3$ & $1024$ & $1$ & SAME & - & $9 \times 9$ & $45 \times 45$ \\
& ConvBlock ($\times 4$) & $1 \times 1$ & $1024$ & $1$ & SAME & - & $9 \times 9$ & $45 \times 45$ \\
\cmidrule{2-9}
& ConvBlock & $3 \times 3$ & $2048$ & $1$ & VALID & - & $7 \times 7$ & $61 \times 61$ \\
& ConvBlock & $3 \times 3$ & $2048$ & $1$ & SAME & - & $7 \times 7$ & $77 \times 77$ \\
& ConvBlock & $1 \times 1$ & $2048$ & $1$ & SAME & - & $7 \times 7$ & $77 \times 77$ \\
\cmidrule{2-9}
& Conv & $1 \times 1$ & $512$ & $1$ & SAME & Leaky ReLU & $7 \times 7$ & $77 \times 77$ \\
& GAP & - & $512$ & - & - & - & $1 \times 1$ & - \\
\midrule
\multirow{7}{*}{\textbf{Location}} & \multirow{2}{*}{Input} & \multirow{2}{*}{-} & \multirow{2}{*}{-} & \multirow{2}{*}{-} & \multirow{2}{*}{-} & - & $5 \times 5 \times 1024$, & \multirow{2}{*}{-} \\
& & & & & & - & $1 \times 1 \times 512$ & \\
\cmidrule{2-9}
& Conv & $1 \times 1$ & $512$ & $1$ & SAME & Leaky ReLU & $5 \times 5 \times 512$ & - \\
& Conv & $1 \times 1$ & $1$ & $1$ & SAME & - & $5 \times 5 \times 1$ & - \\
& $L_2$ Normalization & - & $25$ & - & - & - & $1 \times 25$ & - \\
& Softmax & - & $25$ & - & - & - & $1 \times 25$ & - \\
\midrule
\multirow{4}{*}{\shortstack{\textbf{Positional} \\ \textbf{Encoding}}} & \multirow{2}{*}{Input} & \multirow{2}{*}{-} & \multirow{2}{*}{-} & \multirow{2}{*}{-} & \multirow{2}{*}{-} & - & $1 \times 512$, & \multirow{2}{*}{-} \\
& & & & & & - & $1 \times 512$ & \\
\cmidrule{2-9}
& Fully Connected & - & $512$ & - & - & - & $1 \times 512$ & - \\
\midrule
\multirow{3}{*}{\textbf{Classification}} & Input & - & - & - & - & - & $1 \times 512$ & - \\
\cmidrule{2-9}
& Fully Connected & - & $1000$ & - & - & - & $1 \times 1000$ & - \\

\bottomrule
\end{tabular}
\end{small}
}
\end{center}
\vskip -0.1in
\end{table*}
\setlength{\tabcolsep}{8pt}
\begin{table*}[!t]
\caption{TNet architecture used on fMoW (see Section \ref{sec5_2}).
EfficientNet-B$0$ baseline corresponds to the feature extraction module followed by the classification module.
For simplicity, we provide only the spatial dimensions of the feature extraction module's output.
The location module receives the downsampled output feature map of the $8$-th MBConv block as input (receptive field of $147 \times 147$ px).
The positional encoding module receives a $1 \times 320$ positional encoding vector that is projected to $1 \times 1280$, and then it is added to the second input of the module, which is a $1 \times 1280$ feature vector.}
\label{App_table_3}
\begin{center}
\resizebox{1.0\textwidth}{!}{
\begin{small}
\begin{tabular}{ccccccccc}
\toprule
\multirow{2}{*}{\textbf{Module}} & \textbf{Layer/Block} & \textbf{Kernel} & \multirow{2}{*}{\shortstack{\textbf{\#Output} \\ \textbf{Channels}}} & \multirow{2}{*}{\textbf{Stride}} & \multirow{2}{*}{\shortstack{\textbf{Batch} \\ \textbf{Norm}}} & \multirow{2}{*}{\textbf{Activation}} & \textbf{Output} & \textbf{Receptive} \\
& \textbf{Type} & \textbf{Size} & & & & & \textbf{Size} & \textbf{Field} \\
\midrule
\multirow{21}{*}{\shortstack{\textbf{Feature} \\ \textbf{Extraction}}} & Input & - & - & - & - & - & $224 \times 224$ & - \\
\cmidrule{2-9}
& Conv & $3 \times 3$ & $32$ & $2$ & $\surd$ & SiLU & $112 \times 112$ & $3 \times 3$ \\ 
\cmidrule{2-9}
& MBConv1 & $3 \times 3$ & $16$ & $1$ & - & - & $112 \times 112$ & $7 \times 7$ \\ 
\cmidrule{2-9}
& MBConv6 & $3 \times 3$ & $24$ & $2$ & - & - & $56 \times 56$ & $11 \times 11$ \\ 
& MBConv6 & $3 \times 3$ & $24$ & $1$ & - & - & $56 \times 56$ & $19 \times 19$ \\ 
\cmidrule{2-9}
& MBConv6 & $5 \times 5$ & $40$ & $2$ & - & - & $28 \times 28$ & $35 \times 35$ \\ 
& MBConv6 & $5 \times 5$ & $40$ & $1$ & - & - & $28 \times 28$ & $67 \times 67$ \\ 
\cmidrule{2-9}
& MBConv6 & $3 \times 3$ & $80$ & $2$ & - & - & $14 \times 14$ & $83 \times 83$ \\ 
& MBConv6 & $3 \times 3$ & $80$ & $1$ & - & - & $14 \times 14$ & $115 \times 147$ \\ 
& MBConv6 & $3 \times 3$ & $80$ & $1$ & - & - & $14 \times 14$ & $147 \times 115$ \\ 
\cmidrule{2-9}
& MBConv6 & $5 \times 5$ & $112$ & $1$ & - & - & $14 \times 14$ & $211 \times 211$ \\ 
& MBConv6 ($\times 2$) & $5 \times 5$ & $112$ & $1$ & - & - & $14 \times 14$ & $339 \times 339$ \\ 
\cmidrule{2-9}
& MBConv6 & $5 \times 5$ & $192$ & $2$ & - & - & $7 \times 7$ & $403 \times 403$ \\ 
& MBConv6 ($\times 3$) & $5 \times 5$ & $112$ & $1$ & - & - & $7 \times 7$ & $787 \times 787$ \\ 
\cmidrule{2-9}
& MBConv6 & $3 \times 3$ & $320$ & $1$ & - & - & $7 \times 7$ & $851 \times 851$ \\ 
\cmidrule{2-9}
& Conv & $1 \times 1$ & $1280$ & $1$ & $\surd$ & SiLU & $7 \times 7$ & $851 \times 851$ \\
& GAP & - & $1280$ & - & - & - & $1 \times 1$ & - \\
\midrule
\multirow{8}{*}{\textbf{Location}} & Input & - & - & - & - & - & $3 \times 3 \times 80$ & - \\
\cmidrule{2-9}
& Conv & $1 \times 1$ & $80$ & $1$ & - & SiLU & $3 \times 3 \times 80$ & - \\
& SE-($80, \; 0.5$) & - & $80$ & - & - & - & $3 \times 3 \times 80$ & - \\
& Conv & $1 \times 1$ & $80$ & $1$ & - & SiLU & $3 \times 3 \times 80$ & - \\
& Conv & $1 \times 1$ & $1$ & $1$ & - & - & $3 \times 3 \times 1$ & - \\
& $L_2$ Normalization & - & $9$ & - & - & - & $1 \times 9$ & - \\
& Softmax & - & $9$ & - & - & - & $1 \times 9$ & - \\
\midrule
\multirow{5}{*}{\shortstack{\textbf{Positional} \\ \textbf{Encoding}}} & Input & - & - & - & - & - & $1 \times 320$ & - \\
\cmidrule{2-9}
& Fully Connected & - & $1280$ & - & - & - & $1 \times 1280$ & - \\
\cmidrule{2-9}
& Input & - & - & - & - & - & $1 \times 1280$ & - \\
\cmidrule{2-9}
& Add & - & $1280$ & - & - & SiLU & $1 \times 1280$ & - \\
\midrule
\multirow{3}{*}{\textbf{Classification}} & Input & - & - & - & - & - & $1 \times 1280$ & - \\
\cmidrule{2-9}
& Fully Connected & - & $62$ & - & - & - & $1 \times 62$ & - \\
\bottomrule
\end{tabular}
\end{small}
}
\end{center}
\vskip -0.1in
\end{table*}
We first present the architectures we use in our experiments (see Section \ref{sec5}), and then we provide more details about the design of individual modules.
In Table \ref{App_table_1}, we provide the building blocks of our architectures.

\subsubsection{Models used on ImageNet}
\label{App_sec_Arch_1}
In Table \ref{App_table_2}, we provide the TNet architecture we use in our experiments on ImageNet \cite{deng2009imagenet} (see Section \ref{sec5_1}).
BagNet-$77$ baseline corresponds to TNet's feature extraction module followed by the classification module.

BagNet-$77$ results from BagNet-$77$-lowD with $3$ modifications.
First, we replace "VALID" padding of some convolutional layers with "SAME", to obtain less aggressive reduction of the spatial dimensions;
the base resolution of TNet is $77 \times 77$ px, instead of $224 \times 224$ px which is the input size of the Saccader's backbone.
Second, we remove Batch Normalization due to technical issues in preliminary experiments (Batch Norm was successfully used in later experiments with the other datasets).
Third, we use Leaky ReLU instead of ReLU activations, to allow non-zero gradients for negative inputs.

\subsubsection{Models used on fMoW}
\label{App_sec_Arch_2}
In Table \ref{App_table_3}, we provide the TNet architecture we use in our experiments on fMoW \cite{fmow2018} (see Section \ref{sec5_2}).
EfficientNet-B$0$ baseline corresponds to TNet's feature extraction module followed by the classification module.

\subsubsection{Models used on CUB-$\mathbf{200}$-$\mathbf{2011}$ and NABirds}
\label{App_sec_Arch_3}
\setlength{\tabcolsep}{8pt}
\begin{table*}[!t]
\caption{The feature weighting module of TNet-B$0$ (see Section \ref{secMFT}).
The input consists of the $N$ feature vectors extracted while attending to a sequence of $N-1$ locations;
$N-1$ vectors are extracted from the attended locations, and a feature vector from the downscaled version of the whole image ($1$st processing level).
The module first calculates $N$ weights that sum up to $1$, and then, it uses them to perform a weighted average of the $N$ input feature vectors.}
\label{App_table_4}
\begin{center}
\resizebox{1.0\textwidth}{!}{
\begin{small}
\begin{tabular}{ccccccccc}
\toprule
\multirow{2}{*}{\textbf{Module}} & \textbf{Layer/Block} & \textbf{Kernel} & \multirow{2}{*}{\shortstack{\textbf{\#Output} \\ \textbf{Channels}}} & \multirow{2}{*}{\textbf{Stride}} & \multirow{2}{*}{\textbf{Padding}} & \multirow{2}{*}{\textbf{Activation}} & \textbf{Output} \\
& \textbf{Type} & \textbf{Size} & & & & & \textbf{Size} \\
\midrule
\multirow{5}{*}{\shortstack{\textbf{Feature} \\ \textbf{Weighting}}} & Input & - & - & - & - & - & $1 \times N \times 1280$ \\
\cmidrule{2-8}
& SE-($1280, \; 0.25$) & - & $1280$ & - & - & - & $1 \times N \times 1280$ \\
& Conv & $1 \times 1$ & $1$ & $1$ & SAME & - & $1 \times N \times 1$ \\
& Softmax & - & $N$ & - & - & - & $1 \times N$ \\
& Multiply & - & $1280$ & - & - & - & $1 \times 1280$ \\
\bottomrule
\end{tabular}
\end{small}
}
\end{center}
\vskip -0.1in
\end{table*}

We get the feature extraction module of each TNet-B$i$, $i \in \{0, 1, ..., 4\}$, by removing the last fully connected layer of the corresponding EN-B$i$ model;
this happens in Table \ref{App_table_3} as well, where we get the feature extraction module of TNet from EN-B$0$.
The location and positional encoding modules are implemented as in Table \ref{App_table_3}, with output channels scaled according to the feature extraction module in use.
For the location module, the attention grid is $5 \times 5$, leading to an output of size $1 \times 25$. Also, for different TNet-B$i$ models, the input to the location module may vary in number of channels; more details are provided in Section \ref{App_sec_2_1_1}.
The classification module is a linear layer, as in Table \ref{App_table_3}. The number of output nodes is equal to the number of classes; $200$ for CUB-$200$-$2011$ \cite{wah2011caltech}, and $555$ for NABirds~\cite{van2015building}.

In Table \ref{App_table_4}, we provide the feature weighting module of TNet-B$0$.
The same design is followed for the other TNet-B$i$ models as well, with output channels scaled according to different feature extraction modules.
The input to the feature weighting module is of variable size, as it depends on the number of attended locations.
More details are provided in Section \ref{App_sec_2_1_3}.

\subsubsection{Location module}
\label{App_sec_2_1_1}
In Tables \ref{App_table_2} and \ref{App_table_3}, we provide two different implementations of the location module.
In Table \ref{App_table_2}, location module receives two inputs.
The first one is a feature map of size $5 \times 5 \times 1024$, which originates from an intermediate layer of the feature extraction module. The spatial dimensions of the feature map are equal to the dimensions of the candidate location grid.
Each $1 \times 1 \times 1024$ vector of the feature map, describes the image region within the corresponding grid cell.

To achieve this, we aim for the receptive field of each pixel in the feature map to align with the image region that it is supposed to describe.
In the specific architecture of Table \ref{App_table_2}, we assume a $5 \times 5$ grid of overlapping cells, and an input to the feature extraction module of fixed size $77~\times~77$~px.
Each grid cell occupies $34.375 \%$ of the corresponding input dimension. Based on that, when the $5 \times 5$ grid is superimposed onto the $77 \times 77$ px input, each cell is approximately of size $27 \times 27$~px.

The layer of the feature extraction module with the closest receptive field size, is in the $8$-th ConvBlock with $29 \times 29$ px. However, the effective receptive field size is usually smaller that the actual receptive field size \cite{luo2017understanding}, as a result, we pick the output feature map of the $13$-th ConvBlock with receptive field $45 \times 45$ px.
The spatial dimensions of this feature map are $9 \times 9$, and we need to downsample it to $5 \times 5$ px.
To this end, we calculate the image level coordinates of the receptive field centers of the feature map pixels, and we pick the $25$ of them with receptive fields that better align with the assumed candidate image regions. Based on our previous remarks about the effective receptive field size, we don't consider perfect alignment to be crucial.

The second input to the location module provides contextual information, and it is the output feature vector of the feature extraction module. This vector is of size $1 \times 1 \times 512$, and we concatenate it across the channel dimension at each spatial position of the input feature map, increasing its size to $5 \times 5 \times 1536$.

We pass the combined feature map through two $1 \times 1$ convolutional layers. The first one fuses the features with the concatenated context. The second one projects each fused vector to a logit value, which represents the relative importance of the corresponding candidate location.

We use the same weights to estimate the importance of each candidate location ($1 \times 1$ convolutions).
We don't want to use different sets of weights (e.g., to have $25$ output heads \cite{uzkent2020learning}),
because this doesn't allow information learned in one location to transfer to other locations. Also, less attended locations (e.g., corners) could lead to a partially trained model with erratic behavior.

The downside of $1 \times 1$ convolutions is that they disregard spatial information. 
To mitigate this problem, 
we enrich the input tensor (the one of size $5 \times 5 \times 1536$) with positional information according to \cite{zhao2020exploring}.
In particular, for each spatial location, we calculate horizontal and vertical coordinates in the normalized range $[-1, 1]$.
Then, we use $2$ linear layers (one for each spatial dimension), to map coordinates into a learned range.
The resulting $2$-dimensional vectors are concatenated across the channel dimension, resulting to an input feature map of size $5 \times 5 \times 1538$.
This is the feature map that we actually pass through the $1 \times 1$ convolutional layers.

The estimated logits are reshaped to a $1 \times 25$ vector, which is first normalized to have $L_2$ norm equal to $1$, and then it is passed through a Softmax layer to get the final parameters of the Categorical distribution. The $L_2$ normalization aims to reduce the variance between logits, because we empirically observe that logit values may be too negative, or very close to zero, leading Softmax outputs to be exactly $0$, and thus hindering the backpropagation of gradients.

The architecture of the location module in Table \ref{App_table_3} is conceptually the same, but has some technical differences. In particular, we provide only one input, the output feature map of the $8$-th MBConv block (selected and downsampled according to the process described before). This means that we don't provide the output vector of the feature extraction module as an additional input. The reason is that its size of $1~\times~1280$ results in a parameter-heavy location module, which is antithetical to the very light design of the feature extraction module.

To inject contextual information to the input feature map, we pass it through a squeeze-and-excitation (SE) block \cite{hu2018squeeze}. Other than that, we follow the design principles described before. We use $1~\times~1$ convolutions, we augment the SE output feature map with $2$-dimensional spatial coordinates' vectors, and we $L_2$ normalize the logits.

In TNet-B$i$ models, location module is implemented as in Table \ref{App_table_3} (see Section \ref{App_sec_Arch_3}).
However, the input feature map may originate from different layers of the feature extraction module, to account for receptive field differences between backbone networks.
For TNet-B$0$, the location module receives as input the output of the $6$-th MBConv block, while for TNet-B$1$, TNet-B$2$ and TNet-B$3$, the input comes from the $7$-th MBConv block.
Finally, for TNet-B$4$, the location module receives as input the output of the $8$-th MBConv block

\subsubsection{Positional encoding module}
\label{App_sec_2_1_2}
In Tables \ref{App_table_2} and \ref{App_table_3}, we provide two different implementations of the positional encoding module.
In both cases, the positional encoding module receives two inputs.
The first one is the output feature vector of the feature extraction module. The second input is a vector that encodes positional information about the image region described by the first input. The encoded positional information is 3-dimensional; the first $2$ dimensions correspond to spatial coordinates, and the $3$rd one to scale.

Given a processing level $l$, we assume that a grid is superimposed onto the input image, where its cells correspond to all possible candidate locations of the level. In the example of Fig. \ref{fig_4}, in the $1$st processing level, the assumed grid consists of a single cell. In the second level ($l=2$) the gird is $2 \times 2$, and for $l=3$ the gird is $4 \times 4$.

The spatial coordinates of the grid cells start with $(0,0)$ in the top left corner, and increase linearly with step $1$ both horizontally and vertically. The scale coordinate is equal to $l-1$. Based on this, each candidate image region has a unique positional triplet $(x, y, s)$, where $x$, $y$ are the spatial coordinates and $s$ is the scale.

We use sine and cosine functions of different frequencies to encode positional triplets $(x, y, s)$ according to \cite{vaswani2017attention}. In particular, for positional encodings of size $1 \times N$, we get:
\begin{subequations}
\label{App_eq8}
\begin{align}
&P_{s}(p, \vec{t}) = \sin(p \cdot \Big( \frac{1}{100} \Big)^{\frac{\vec{t}}{\lfloor \sfrac{N}{6} \rfloor}} ), \nonumber \\
&P_{c}(p, \vec{t}) = \cos(p \cdot \Big( \frac{1}{100} \Big)^{\frac{\vec{t}}{\lfloor \sfrac{N}{6} \rfloor}} ), \nonumber \\
&p \in [x, y, s] \nonumber \\
&\vec{t} = [0, 1, 2, ... \lfloor \sfrac{N}{6} \rfloor] \nonumber
\end{align}
\end{subequations}

The final positional encoding for triplet $(x, y, s)$, results by concatenating $P_{s}(x, \vec{t})$, $P_{c}(x, \vec{t})$, $P_{s}(y, \vec{t})$, $P_{c}(y, \vec{t})$, $P_{s}(s, \vec{t})$ and $P_{c}(s, \vec{t})$.

The main reason we use these positional encodings (instead of, e.g., learned positional embeddings \cite{dosovitskiy2020image}), is that they can generalize to scales and spatial dimensions of arbitrary size.
This is particularly useful for our model, because it has the potential to extend its processing to an arbitrary number of levels.

In Table \ref{App_table_2}, the positional encoding module concatenates its $2$ input vectors, and fuses their information through a linear layer.
In Table \ref{App_table_3}, we implement the positional encoding module differently, because we aim for a relatively smaller number of parameters.
To this end, we use positional encodings of $4$ times smaller dimensionality compared to the input feature vector ($320$ instead of $1280$).
In addition, only the positional encoding is processed by a trainable linear layer.
This linear projection brings the positional encoding to the size of the feature vector, while it provides a learned component to the process of encoding position.
The projected positional encoding is simply added to the input feature vector, and the outcome is passed through a non-linearity.

\subsubsection{Feature weighting module}
\label{App_sec_2_1_3}
Given a sequence of $N-1$ attended locations, TNet extracts a total number of $N$ feature vectors; $N-1$ vectors from the attended locations, and a feature vector from the downscaled version of the whole image ($1$st processing level).
Feature weighting module estimates $N$ weights that sum up to $1$, in order to perform a weighted average of the $N$ feature vectors.

As we can see in Table \ref{App_table_4}, feature weighting module receives $N$ feature vectors as input.
Since the number of attended locations varies, the input to the feature weighting module is of variable size as well.
In order to calculate the $N$ weights, we first inject contextual information to the $N$ feature vectors through a squeeze-and-excitation block. Then, each of the resulting $N$ vectors, is projected to a logit value through a $1 \times 1$ convolutional layer. The $N$ logits are passed through a Softmax layer to get the final weights. The weighted average is implemented via multiplication with the input feature vectors.

\subsection{Training}
\label{App_sec_2_2}

\subsubsection{Training on ImageNet}
\label{App_sec_2_2_1}
To train TNet, we use a single sample ($M = 1$) for the Monte Carlo estimators, and we set $\lambda_f=0.1$ (Eq. \ref{App_eq6}).
We experimented with different $M$ values, e.g., of 2 and 4, but we observed no significant differences in performance. Since the value of $M$ has a multiplicative effect on the batch size, which leads to considerable increase in training time, we set $M = 1$ in all our experiments.
The BagNet-$77$ baseline is trained by minimizing the cross-entropy classification loss.

For both models we use batches of $64$ images, distributed in $4$ GPUs.
We use the Adam optimizer with the default values of $\beta_1 = 0.9$, $\beta_2 = 0.999$ and $\epsilon = 10^{-8}$.
We use xavier initialization \cite{glorot2010understanding} for the weights, and zero initialization for the biases.
For regularization purposes, we use data augmentation that is very similar to the one used in \cite{szegedy2015going}. In particular, given a training image, we get a random crop that covers at least $85 \%$ of the image area, while it has an aspect ratio between $0.5$ and $2.0$. Since we provide inputs of fixed size to our networks ($224 \times 224$~px), we resize the image crops accordingly. Resizing is performed by randomly selecting between $8$ different methods, which include bilinear, nearest neighbor, bicubic, and area interpolation. Also, we randomly flip the resized image crops horizontally, and we apply photometric distortions \cite{howard2013some}. The final image values are scaled in range $[-1, 1]$. Finally, the dropout mentioned in Section \ref{sec5_1}, is spatial \cite{tompson2015efficient}.

Since per-feature regularization plays a crucial role in the performance of TNet, we experimented with a variety of different values for $\lambda_r$ and $\lambda_c$, including $0.1$, $0.3$, $0.5$, $0.7$ and $0.9$, while $\lambda_r$ and $\lambda_c$ were not always set to be equal.
We conducted similar tuning for $\lambda_f$, observing that differences in its value didn’t have the impact that those of $\lambda_r$ and $\lambda_c$ had.
In the following Sections we report only our final choices for the values of $\lambda_r$, $\lambda_c$ and $\lambda_f$, which led to the best performance.

\subsubsection{Training on fMoW}
\label{App_sec_2_2_2}
We first train TNet with inputs of size $448 \times 448$ px, allowing $2$ processing levels.
We train for $40$ epochs with batches of $64$ images (on $4$ GPUs), with initial learning rate of $0.001$ that drops once by a factor of $0.1$.
We use the Adam optimizer with its default parameter values, and we follow the weight initialization of \cite{tan2019efficientnet}.

We attend to a fixed number of $2$ locations. We use $\lambda_f = 0.1$ and per-feature regularization with $\lambda_c~=~\lambda_r~=~0.2$.
We use a single sample for the Monte Carlo estimators.

We use dropout before the linear layer of the classification module with $0.5$ drop probability.
We use stochastic depth \cite{huang2016deep} with drop probability that increases linearly to a maximum value of $0.3$.
We use the data augmentation technique described in Section \ref{App_sec_2_2_1}.

We fine-tune TNet for $10$ epochs on images of $896~\times~896$~px, with a fixed number of $2$ attended location in the $2$nd processing level, and $1$ in the $3$rd ($4$ in total). Compared to the previous step, we increase the maximum drop probability of stochastic depth to $0.5$, and we set $\lambda_c~=~\lambda_r~=~0.05$.
Also, we only use features extracted until the $2$nd processing level in per-feature regularization (features extracted in the $3$rd processing level are excluded).

We use different input images to train $4$ EfficientNet-B$0$ baselines. For the first baseline we use images cropped according to the bounding box annotations, and resized to $224~\times~224$~px. We train for $65$ epochs with batches of $64$ images, on $4$ GPUs. Our initial learning rate is $0.001$, and it drops once by a factor of $0.1$.
We use the Adam optimizer with its default parameter values, and we follow the weight initialization of \cite{tan2019efficientnet}.

We use dropout before the final classification layer with $0.75$ drop probability, and $L_2$ regularization with weight of $10^{-5}$.
We use stochastic depth with drop probability that increases linearly to a maximum value of $0.5$.
We use the data augmentation technique described in Section \ref{App_sec_2_2_1}.

The second baseline is trained on the original images, resized to $224~\times~224$~px. The only difference with the training of the previous baseline is that we train for $60$ epochs.

The third baseline is trained on the original images resized to $448~\times~448$~px. We train for $30$ epochs with batches of $32$ images. We reduce stochastic depth maximum drop probability to $0.3$. All other training hyperparameters remain the same.

The fourth baseline is trained on the original images resized to $896~\times~896$~px. We train for $30$ epochs with batches of $32$ images. We set dropout drop probability to $0.3$, and stochastic depth maximum drop probability to $0.2$. All other training hyperparameters remain the same.

\subsubsection{Training on CUB-$\mathbf{200}$-$\mathbf{2011}$ and NABirds}
\label{App_sec_2_2_3}
CUB-$200$-$2011$ \cite{wah2011caltech} and NABirds \cite{van2015building} are fine-grained classification datasets with images of different bird species.
Images from different classes may exhibit very high visual similarity, and as a result, successful classification requires learning subtle discriminative features.
To alleviate this problem, we consider the contrastive loss term from \cite{he2021transfg}:
\begin{equation}
\label{App_eq10}
L_{con} = \lambda_{con} \cdot \frac{1}{N^2} \sum_{i}^{N} \Big[ \sum_{j: y_i=y_j}^{N} \big( 1 - \text{cos\_sim}(f_i, f_j) \big) + \sum_{j: y_i \neq y_j}^{N} \text{max} \big( \text{cos\_sim}(f_i. f_j) - \alpha, 0 \big) \Big]
\end{equation}

where $N$ is the batch size, $x_k$ is the $k$-th image in the batch, $f_k$ is a feature vector representing $x_k$, $y_k$ is the class label of $x_k$, and $\text{cos\_sim}(\cdot, \cdot)$ is a function that receives two vectors as input and calculates their cosine similarity.
$\alpha$ is a hyperparameter that constitutes a similarity threshold.
$\lambda_{con}$ is a hyperparameter that specifies the relative importance of $L_{con}$ within the total loss used for training.

The first term in Equation \ref{App_eq10} is used to maximize the similarity between feature vectors that represent images from the same class. The second term is used to not allow the similarity between feature vectors that represent images from different class to exceed $\alpha$.

For all models, we use pre-trained weights that are available in \cite{efficientnet2021github}. In particular, we use the weights of EfficientNet models trained with NoisyStudent \cite{xie2020self} and RandAugment \cite{cubuk2020randaugment} on ImageNet with extra JFT-$300$M unlabeled data.

\textbf{Training on CUB-$\mathbf{200}$-$\mathbf{2011}$.} We train TNet-B$0$ on images of size $448 \times 448$ px, for $200$ epochs, with batches of $64$ images, on $4$ NVIDIA Quadro RTX $8000$ GPUs. The feature extraction module is initialized with pre-trained weights, while for the rest of the modules we follow the random weight initialization of \cite{tan2019efficientnet}. For the weights of the feature extraction module we use a learning rate of $10^{-4}$, while for the rest of the weights we use a learning rate of $10^{-3}$. Both learning rates drop once by a factor of $0.1$.
We use the Adam optimizer with its default parameter values.

We attend to a fixed number of $5$ locations, with processing extended to $2$ levels.
We use the learning rule of Eq. \ref{eq2}, with $\lambda_f = 0.1$, and $\lambda_c~=~\lambda_r~=~0.3$.
We use a single sample for the Monte Carlo estimators.

We use dropout \cite{srivastava2014dropout} before the linear layer of the classification module with $0.75$ drop probability.
We use stochastic depth \cite{huang2016deep} with drop probability that increases linearly to a maximum value of $0.5$.
We use $L_2$ regularization with weight of $10^{-4}$.
We use contrastive loss with $\lambda_{con} = 100$ and $\alpha = 0.4$.

For data augmentation, given a training image, we get a random crop that covers at least $85 \%$ of the image area, while it has an aspect ratio between $0.5$ and $2.0$.
We resize the image crop to $448 \times 448$ px by randomly selecting between $8$ different resizing methods.
We randomly flip the resized image crops horizontally.
We don't apply photometric distortions because color is a discriminative feature for bird species.
We perform random translation and rotation of the image.
The final image values are scaled in the range of $[-1, 1]$.

We train TNet-B$1$, TNet-B$2$ and TNet-B$3$ by following the same training procedure we described for TNet-B$0$.
For TNet-B$4$, the only differences is that we train for fewer epochs, $125$ instead of $200$.

We train all EN-B$i$, $i \in \{0, 2, ..., 4\}$ baselines under the same training regime.
We train on images of size $448 \times 448$ px, for $200$ epochs, with batches of $64$ images, on $4$ NVIDIA Quadro RTX $8000$ GPUs.
We initialize all layers with pre-trained weights, except the last fully connected layer, which size depends on the number of output classes. This last output layer is randomly initialized according to \cite{tan2019efficientnet}.

For layers initialized with pre-trained weights, we use a learning rate of $5 \cdot 10^{-5}$, while for the output layer we use a learning rate of $10^{-3}$. Both learning rates drop once by a factor of $0.1$.
We use the cross entropy loss, and we add a contrastive loss term with $\lambda_{con} = 100$ and $\alpha = 0.4$.
We use Adam optimizer with its default parameter values.
We use the regularization and data augmentation methods we described for the TNet-B$i$ models, with the same hyper-parameters as well.

\textbf{Training on NABirds.} We train all TNet-B$i$, $i \in \{0, 2, ..., 4\}$ models according to the procedure we followed on CUB-$200$-$2011$, and we only change some hyperparameter values. In particular, we train for $100$ epochs, and we set $\lambda_{con} = 50$ for the contrastive loss term.
We train TNet-B$0$, TNet-B$1$ and TNet-B$2$ with a fixed number of $5$ attended locations, while we train TNet-B$3$ and TNet-B$1$ with $3$ attended locations.

We train all EN-B$i$, $i \in \{0, 2, ..., 4\}$ models according to the procedure we followed on CUB-$200$-$2011$.
The only difference is that we train for $100$ epochs, and we set $\lambda_{con} = 50$ for the contrastive loss term.

\setlength{\tabcolsep}{8pt}
\begin{table*}[!t]
\caption{Detailed results on CUB-$200$-$2011$ dataset \cite{wah2011caltech} (see Section \ref{secMFT}).}
\label{App_table_5}
\begin{center}
\begin{small}
\begin{tabular}{cccccc}
\toprule
\textbf{Model} & \textbf{\# Locs} & \textbf{Top-}$\mathbf{1}$ \textbf{Acc.}  & \textbf{Top-}$\mathbf{5}$ \textbf{Acc.} & \textbf{FLOPs (B)} & \textbf{Params (M)} \\
\midrule
\textbf{EfficientNet-B}$\mathbf{0}$ & - & $86.49 \%$ & $96.82 \%$ & $1.55$ & $4.31$ \\
\textbf{EfficientNet-B}$\mathbf{1}$ & - & $88.25 \%$ & $97.55 \%$ & $2.29$ & $6.83$ \\
\textbf{EfficientNet-B}$\mathbf{2}$ & - & $88.13 \%$ & $97.34 \%$ & $2.65$ & $8.05$ \\
\textbf{EfficientNet-B}$\mathbf{3}$ & - & $88.42 \%$ & $97.38 \%$ & $3.88$ & $11.09$ \\
\textbf{EfficientNet-B}$\mathbf{4}$ & - & $89.08 \%$ & $97.26 \%$ & $6.09$ & $18.03$ \\
\midrule
\textbf{ResNet-}$\mathbf{50}$ & - & $84.5 \%$ & - & $16.35$ & $23.99$ \\
\textbf{API-Net} \cite{zhuang2020learning} & - & $90.0 \%$ & - & - & $29$ \\
\textbf{TransFG} \cite{he2021transfg} & - & $91.7 \%$ & - & - & $86$ \\
\midrule
\multirow{6}{*}{\textbf{TNet-B}$\mathbf{0}$} & $5$ & $87.75 \%$ & $97.27 \%$ & $2.32$ & \multirow{6}{*}{$5.56$} \\
& $4$ & $87.59 \%$ & $97.26 \%$ & $1.94$ & \\
& $3$ & $87.66 \%$ & $97.26 \%$ & $1.55$ & \\
& $2$ & $87.07 \%$ & $97.17 \%$ & $1.16$ & \\
& $1$ & $85.66 \%$ & $96.5 \%$ & $0.78$ & \\
& $0$ & $77.84 \%$ & $93.5 \%$ & $0.39$ & \\
\midrule
\multirow{6}{*}{\textbf{TNet-B}$\mathbf{1}$} & $5$ & $88.33 \%$ & $97.67 \%$ & $3.44$ & \multirow{6}{*}{$8.07$} \\
& $4$ & $88.18 \%$ & $97.55 \%$ & $2.86$ & \\
& $3$ & $88.35 \%$ & $97.46 \%$ & $2.29$ & \\
& $2$ & $87.56 \%$ & $97.20 \%$ & $1.72$ & \\
& $1$ & $85.93 \%$ & $96.63 \%$ & $1.15$ & \\
& $0$ & $79.72 \%$ & $94.56 \%$ & $0.57$ & \\
\midrule
\multirow{6}{*}{\textbf{TNet-B}$\mathbf{2}$} & $5$ & $88.35 \%$ & $97.60 \%$ & $3.99$ & \multirow{6}{*}{$9.55$} \\
& $4$ & $88.20 \%$ & $97.48 \%$ & $3.32$ & \\
& $3$ & $87.80 \%$ & $97.27 \%$ & $2.66$ & \\
& $2$ & $87.64 \%$ & $97.20 \%$ & $1.99$ & \\
& $1$ & $86.16 \%$ & $96.81 \%$ & $1.33$ & \\
& $0$ & $80.10 \%$ & $94.46 \%$ & $0.67$ & \\
\midrule
\multirow{6}{*}{\textbf{TNet-B}$\mathbf{3}$} & $5$ & $89.35 \%$ & $97.88 \%$ & $5.84$ & \multirow{6}{*}{$12.87$} \\
& $4$ & $89.02 \%$ & $97.74 \%$ & $4.86$ & \\
& $3$ & $89.1 \%$ & $97.67 \%$ & $3.89$ & \\
& $2$ & $88.44 \%$ & $97.50 \%$ & $2.92$ & \\
& $1$ & $87.18 \%$ & $97.24 \%$ & $1.95$ & \\
& $0$ & $81.64 \%$ & $95.41 \%$ & $0.97$ & \\
\midrule
\multirow{6}{*}{\textbf{TNet-B}$\mathbf{4}$} & $5$ & $90.06 \%$ & $98.29 \%$ & $9.15$ & \multirow{6}{*}{$20.46$} \\
& $4$ & $89.97 \%$ & $98.21 \%$ & $7.63$ & \\
& $3$ & $89.92 \%$ & $98.1 \%$ & $6.1$ & \\
& $2$ & $89.27 \%$ & $98.02 \%$ & $4.58$ & \\
& $1$ & $87.95 \%$ & $97.39 \%$ & $3.05$ & \\
& $0$ & $82.59 \%$ & $95.88 \%$ & $1.53$ & \\
\bottomrule
\end{tabular}
\end{small}
\end{center}
\vskip -0.1in
\end{table*}

\setlength{\tabcolsep}{8pt}
\begin{table*}[!t]
\caption{Detailed results on NABirds dataset \cite{van2015building} (see Section \ref{secMFT}).}
\label{App_table_6}
\begin{center}
\begin{small}
\begin{tabular}{cccccc}
\toprule
\textbf{Model} & \textbf{\# Locs} & \textbf{Top-}$\mathbf{1}$ \textbf{Acc.}  & \textbf{Top-}$\mathbf{5}$ \textbf{Acc.} & \textbf{FLOPs (B)} & \textbf{Params (M)} \\
\midrule
\textbf{EfficientNet-B}$\mathbf{0}$ & - & $84.97 \%$ & $96.77 \%$ & $1.55$ & $4.76$ \\
\textbf{EfficientNet-B}$\mathbf{1}$ & - & $86.55 \%$ & $97.44 \%$ & $2.29$ & $7.29$ \\
\textbf{EfficientNet-B}$\mathbf{2}$ & - & $86.79 \%$ & $97.50 \%$ & $2.65$ & $8.55$ \\
\textbf{EfficientNet-B}$\mathbf{3}$ & - & $87.63 \%$ & $97.50 \%$ & $3.88$ & $11.63$ \\
\textbf{EfficientNet-B}$\mathbf{4}$ & - & $87.87 \%$ & $97.59 \%$ & $6.09$ & $18.67$ \\
\midrule
\textbf{API-Net} \cite{zhuang2020learning} & - & $88.1 \%$ & - & - & $29$ \\
\textbf{TransFG} \cite{he2021transfg} & - & $90.8 \%$ & - & - & $86$ \\
\midrule
\multirow{6}{*}{\textbf{TNet-B}$\mathbf{0}$} & $5$ & $86.56 \%$ & $97.78 \%$ & $2.33$ & \multirow{6}{*}{$6.01$} \\
& $4$ & $86.49 \%$ & $97.72 \%$ & $1.94$ & \\
& $3$ & $86.16 \%$ & $97.57 \%$ & $1.55$ & \\
& $2$ & $85.56 \%$ & $97.22 \%$ & $1.16$ & \\
& $1$ & $83.87 \%$ & $96.45 \%$ & $0.78$ & \\
& $0$ & $73.82 \%$ & $91.95 \%$ & $0.39$ & \\
\midrule
\multirow{6}{*}{\textbf{TNet-B}$\mathbf{1}$} & $5$ & $87.85 \%$ & $98.15 \%$ & $3.44$ & \multirow{6}{*}{$8.52$} \\
& $4$ & $87.63 \%$ & $98.06 \%$ & $2.86$ & \\
& $3$ & $87.20 \%$ & $97.89 \%$ & $2.29$ & \\
& $2$ & $86.39 \%$ & $97.57 \%$ & $1.72$ & \\
& $1$ & $84.65 \%$ & $96.77 \%$ & $1.15$ & \\
& $0$ & $76.99 \%$ & $93.61 \%$ & $0.57$ & \\
\midrule
\multirow{6}{*}{\textbf{TNet-B}$\mathbf{2}$} & $5$ & $87.52 \%$ & $97.92 \%$ & $3.99$ & \multirow{6}{*}{$10.05$} \\
& $4$ & $87.22 \%$ & $97.81 \%$ & $3.32$ & \\
& $3$ & $86.73 \%$ & $97.56 \%$ & $2.66$ & \\
& $2$ & $85.84 \%$ & $97.15 \%$ & $1.99$ & \\
& $1$ & $83.93 \%$ & $96.35 \%$ & $1.33$ & \\
& $0$ & $76.58 \%$ & $93.15 \%$ & $0.67$ & \\
\midrule
\multirow{6}{*}{\textbf{TNet-B}$\mathbf{3}$} & $5$ & $88.33 \%$ & $98.06 \%$ & $5.84$ & \multirow{6}{*}{$13.42$} \\
& $4$ & $88.26 \%$ & $98.00 \%$ & $4.87$ & \\
& $3$ & $87.98 \%$ & $97.78 \%$ & $3.89$ & \\
& $2$ & $87.47 \%$ & $97.56 \%$ & $2.92$ & \\
& $1$ & $86.01 \%$ & $96.98 \%$ & $1.95$ & \\
& $0$ & $78.78 \%$ & $93.74 \%$ & $0.97$ & \\
\midrule
\multirow{6}{*}{\textbf{TNet-B}$\mathbf{4}$} & $5$ & $88.41 \%$ & $98.04 \%$ & $9.15$ & \multirow{6}{*}{$21.09$} \\
& $4$ & $88.25 \%$ & $97.95 \%$ & $7.63$ & \\
& $3$ & $88.07 \%$ & $97.76 \%$ & $6.1$ & \\
& $2$ & $87.53 \%$ & $97.48 \%$ & $4.58$ & \\
& $1$ & $86.12 \%$ & $96.82 \%$ & $3.05$ & \\
& $0$ & $79.57 \%$ & $93.61 \%$ & $1.53$ & \\
\bottomrule
\end{tabular}
\end{small}
\end{center}
\vskip -0.1in
\end{table*}

\subsection{Metrics}
\label{App_sec_3}
We calculate the FLOPs of a convolutional layer in the following way:
\begin{equation}
\label{App_eq9}
N_{FLOPs} = (C_{in} \cdot k^2) \cdot (H_{out} \cdot W_{out} \cdot C_{out})
\end{equation}

where $C_{in}$ is the number of channels in the input feature map, $k \times k$ are the spatial dimensions of the convolutional kernel, $H_{out} \times W_{out}$ is the spatial resolution of the output, and $C_{out}$ is the number of output channels.
Each time the kernel is applied, we make $C_{in} \cdot k^2$ multiplications, and we apply the kernel $H_{out} \cdot W_{out} \cdot C_{out}$ times (number of output pixels).
For fully connected layers, simply holds $k = 1$ and $H_{out}~=~W_{out}~=~1$.

Equation \ref{App_eq9} accounts only for multiplications. If we consider additions as well, the number of FLOPs approximately doubles.
We use Eq. \ref{App_eq9} because it allows us to calculate FLOPs for our EfficientNet baselines that are in accordance with the FLOPs reported in \cite{tan2019efficientnet}.

We time our models during inference by using $45$ sets of $10$ batches with $64$ images in each batch. For each model, we calculate the average value and the standard deviation among the $45$ sets of batches. These are the time measurements reported in Tables \ref{table_1} and \ref{table_2}.
We measure memory requirements in batches of $64$ images, by using the TensorFlow memory profiler.
During profiling, we disregard the first processing iterations, to avoid any computational and memory overhead that stems from the creation of the TensorFlow graph.
Finally, TensorFlow automatically calculates the number of our models' parameters.

\subsection{Results}
\label{App_sec_4}
\subsubsection{Results on fMoW}
\label{App_sec_4_1}
\begin{figure*}[!t]
\centering
\includegraphics[width=1.0\textwidth]{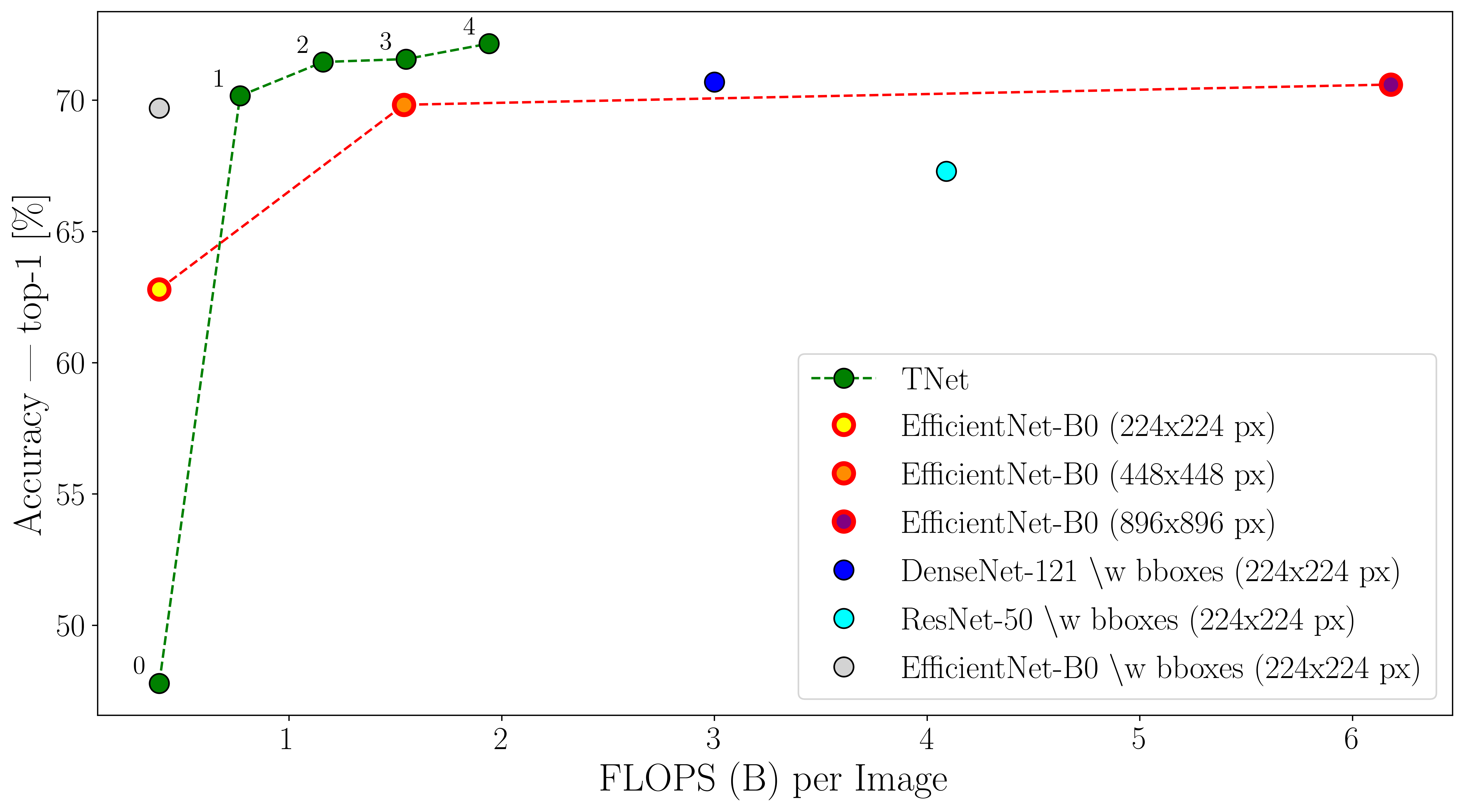}
\caption{Graphical representation of the main results on fMoW (see Table \ref{table_2}).
Numeric annotations correspond to the number of attended locations.
}
\label{App_fig_fmow_plot}
\end{figure*}
In Figure \ref{App_fig_fmow_plot} we plot the main results from Table \ref{table_2}.

\subsubsection{Results on CUB-$\mathbf{200}$-$\mathbf{2011}$ and NABirds}
\label{App_sec_4_2}
We present our results on CUB-$200$-$2011$ \cite{wah2011caltech} and NABirds \cite{van2015building} datasets in Tables \ref{App_table_5} and \ref{App_table_6} respectively.
API-Net \cite{zhuang2020learning} is using a DenseNet-$161$ backbone \cite{huang2017densely}, and TransFG~\cite{he2021transfg} is using a ViT-B/$16$~\cite{dosovitskiy2020image}.

\section{Attention policy and interpretability}
\label{App_sec_att}
\subsection{Quantitative analysis}
\label{App_sec_att_1}
\setlength{\tabcolsep}{8pt}
\begin{table}[!t]
\caption{Precision, recall, and image coverage, calculated on ImageNet and fMoW.
We use bounding boxes that are available for $544,546$ training images from ImageNet, and bounding boxes that are available for all testing images of fMoW.
Attended locations originate from the $2$nd processing level of TNet.
Precision on ImageNet is high (small part of the background area is attended), and recall is low (small part of the object of interest is attended); the opposite is observed on fMoW.
We attribute this behavior to the bigger size of objects of interest in ImageNet compared to fMoW, and to the smaller attention grid cells that we use on ImageNet.
Coverage does not increase linearly with locations' number, because of their overlap.
Attending to $\le 25 \%$ of image area suffices to outperform almost all baselines in Tables \ref{table_1} and \ref{table_2}.
}
\label{table_PRC}
\begin{center}
\begin{small}
\renewcommand{\arraystretch}{1.2}
\begin{tabular}{lccccc}
\toprule
\textbf{Dataset} & \textbf{\# Locs} & \textbf{Precision} & \textbf{Recall} & \textbf{Coverage} &  \textbf{Top-$\mathbf{1}$ Acc.} \\
\midrule
\multirow{5}{*}{\textbf{ImageNet}} & $1$ & $75.52 \%$ & $25.62 \%$ & $11.71 \%$ & $73.12 \%$ \\
& $2$ & $72.9 \%$ & $38.72 \%$ & $19.65 \%$ & $74.12 \%$ \\
& $3$ & $70.22 \%$ & $47.49 \%$ & $26.35 \%$ & $74.41 \%$ \\
& $4$ & $67.66 \%$ & $54.84 \%$ & $32.46 \%$ & $74.58 \%$ \\
& $5$ & $65.3 \%$ & $60.7 \%$ & $38.21 \%$ & $74.62 \%$ \\
\midrule
\multirow{2}{*}{\textbf{fMoW}} & $1$ & $31.92 \%$ & $86.51 \%$ & $25.0 \%$ & $70.17 \%$ \\
& $2$ & $24.37 \%$ & $90.06 \%$ & $37.91 \%$ & $71.46 \%$ \\
\bottomrule
\end{tabular}
\end{small}
\end{center}
\end{table}

We quantify the localization capabilities of TNet, by using bounding box annotations that are available for ImageNet and fMoW.
In particular, given an image and a bounding box, we use the attended image regions at the $2$nd processing level, in order to compute precision and recall in the following way:
\begin{align}
\label{prec}
precision &= \frac{|S_{att} \cap S_{bbox}|}{|S_{att}|}, \; S_{att} \neq \emptyset \\
\label{rec}
recall &= \frac{|S_{att} \cap S_{bbox}|}{|S_{bbox}|}, \; S_{bbox} \neq \emptyset
\end{align}

where $S_{att}$ is the set of pixels that belong to image regions attended by the location module, $S_{bbox}$ is the set of pixels that belong to the bounding box of the object of interest, and $|S|$ denotes the cardinality of a set $S$.
We assume $S_{att} \neq \emptyset$, and $S_{bbox} \neq \emptyset$, meaning that for every image, we attend to at least $1$ location, and we have a bounding box with area greater than $0$.
Both precision and recall take values within $[0, 1]$.
Precision measures the percentage of the attended regions' area that overlaps with the bounding box.
When precision gets smaller, more background (image area outside the bounding box) is attended.
Recall measures the percentage of the bounding box area that is attended.
When recall gets smaller, a smaller part from the object of interest (image area inside the bounding box) is attended.

In Table \ref{table_PRC}, we calculate precision and recall by using bounding boxes that are available for $544,546$ training images on ImageNet, and bounding boxes that are available for every image in the test set of fMoW.
Coverage corresponds to the percentage of the image area that is covered by attended locations.
Attended locations usually overlap, and as a result, coverage does not increase linearly with the number of locations. The fact that we use images from the training set of ImageNet, could potentially lead to biased results in Table \ref{table_PRC}.
In an attempt to test this, we calculate coverage on the validation set of ImageNet, since it doesn't require bounding box annotations.
We find that coverage values are almost identical to the ones reported in Table \ref{table_PRC}.

We observe that precision on ImageNet is high (small part of the background area is attended), while recall is low (small part of the object of interest is attended).
This means that attended locations mainly occupy a limited area within the bounding boxes of the objects of interest.
The opposite is observed in fMoW, with low precision, and high recall.
We primarily attribute this difference in behavior to two factors.
First, bounding boxes from ImageNet are bigger on average compared to those from fMoW.
In particular, on average, a bounding box from ImageNet covers $46 \%$ of the image area, while the same metric is $14 \%$ for bounding boxes from fMoW.
Second, the attention grid cells are smaller on ImageNet ($1$ location has coverage $11.71 \%$ on ImageNet, and $25 \%$ on fMoW), and as a result, the attention policy can be more precise.

The fact that bounding boxes from fMoW have an image coverage of only $14 \%$, can be used to explain the drop in accuracy that is observed in Table \ref{table_2}, when TNet extends processing from $4$ to $6$ locations.
Since objects of interest are small and successfully located (high recall), attending to more locations is expected to mostly add uninformative background clutter.

\subsection{Qualitative examples}
\label{App_sec_att_2}
\begin{figure*}[!t]
\centering
\includegraphics[width=1.0\textwidth]{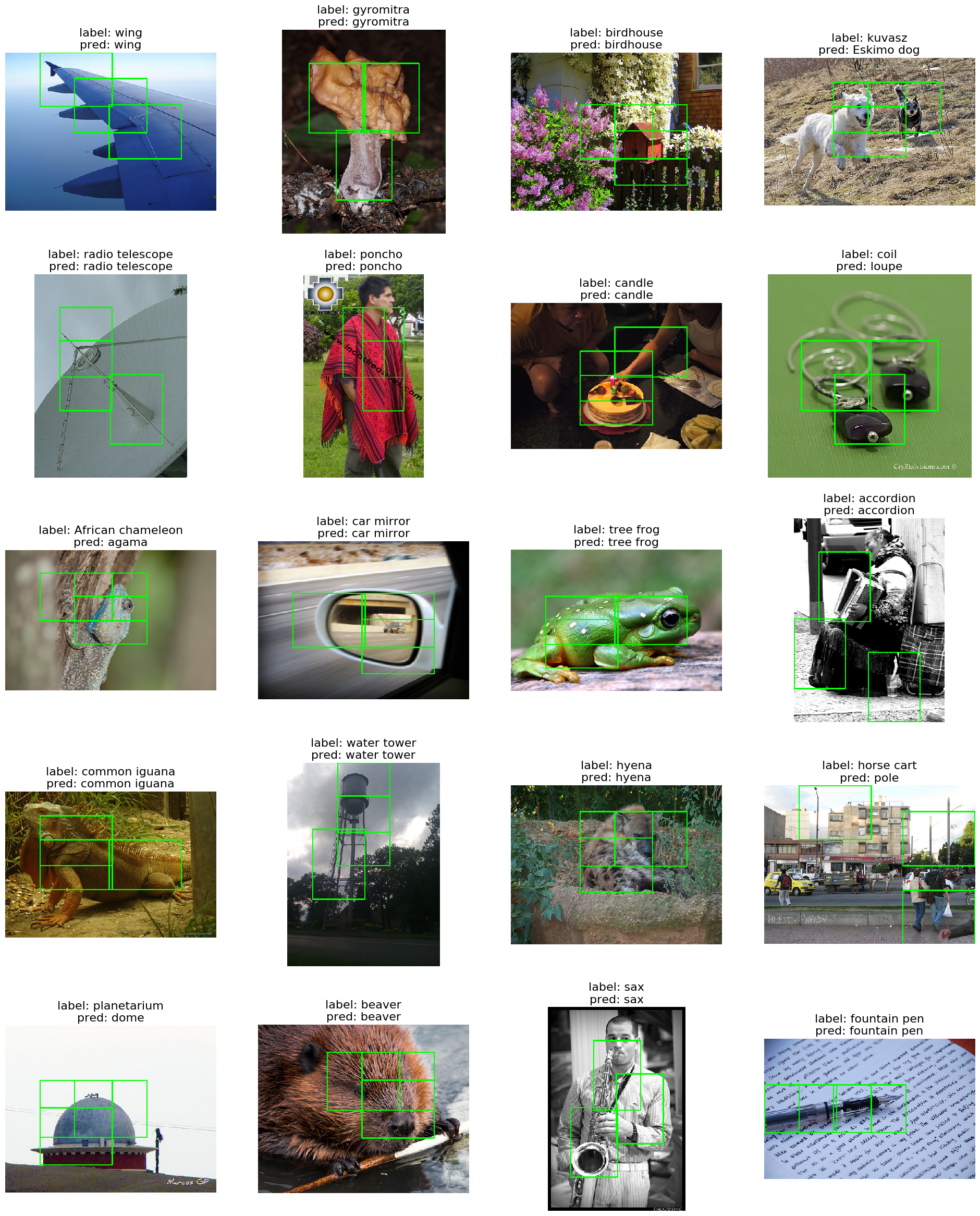}
\caption{Attention policy examples with $3$ locations on the ImageNet validation set.
For every image, the correct and predicted labels are provided.
} 
\label{App_fig_1}
\end{figure*}

\begin{figure*}[!t]
\centering
\includegraphics[width=1.0\textwidth]{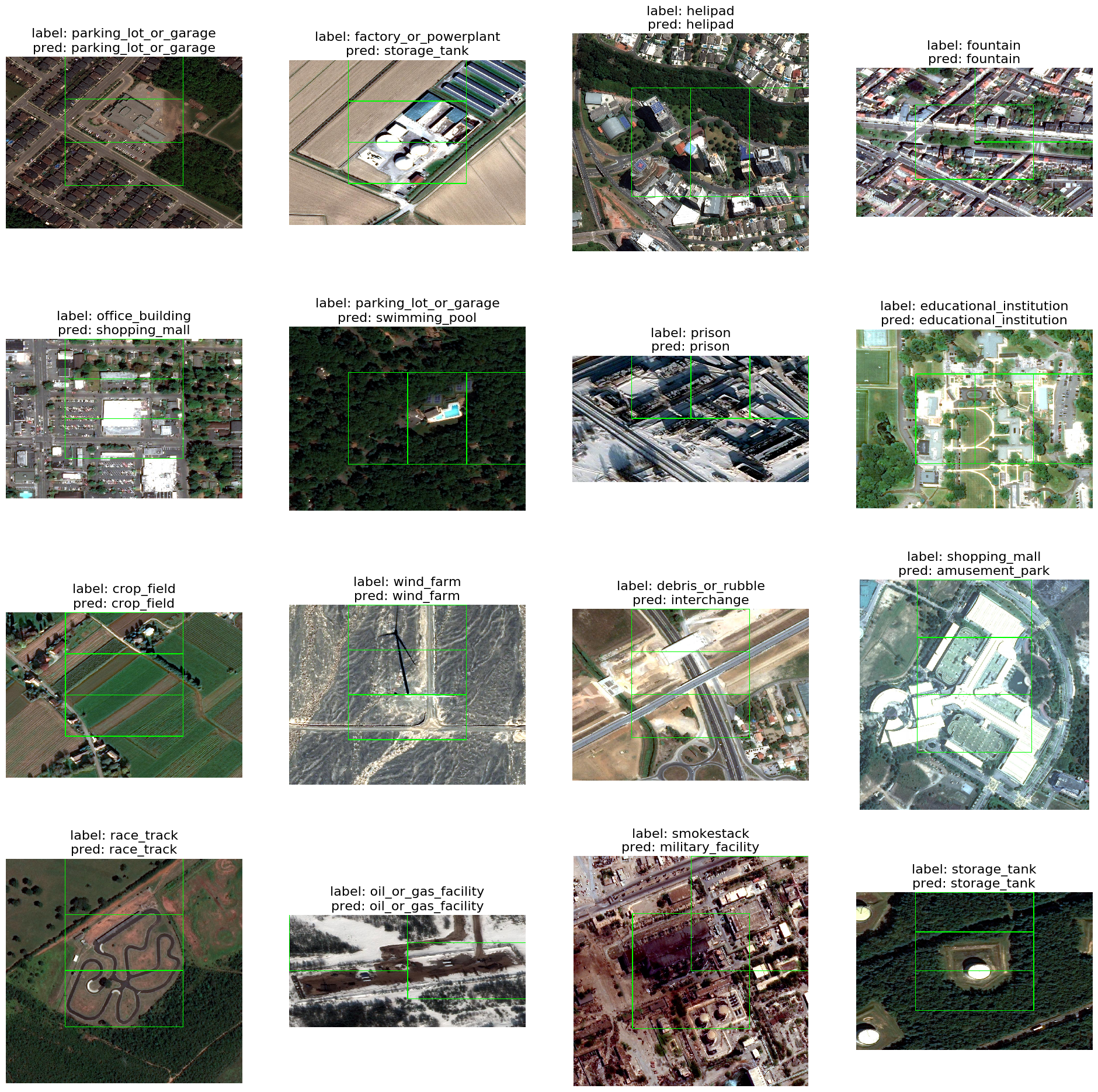}
\caption{Attention policy examples with $2$ locations ($2$nd processing level) on the fMoW test set.
For every image, the correct and predicted labels are provided.
}
\label{App_fig_2}
\end{figure*}

\begin{figure*}[!t]
\centering
\includegraphics[width=0.95\textwidth]{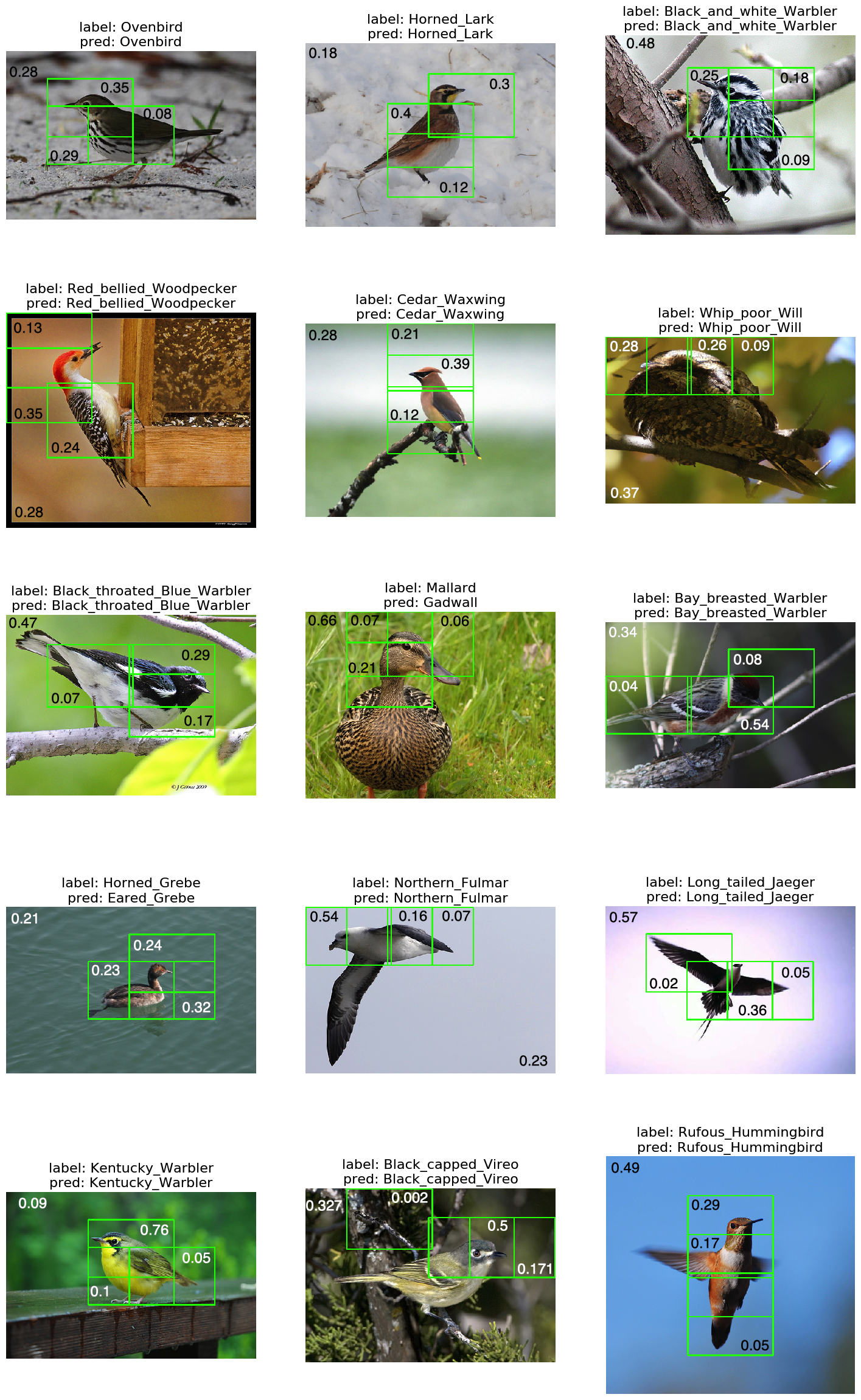}
\caption{Examples from attention policy learned on CUB-$200$-$2011$. Numeric annotations correspond to weights predicted by the feature weighting module, for the top $3$ locations and the downscaled version of the whole image ($1$st processing level).
Weights sum up to $1$.
For every image, the correct and predicted labels are provided.
}
\label{App_fig_3}
\end{figure*}

\begin{figure*}[!t]
\centering
\includegraphics[width=0.95\textwidth]{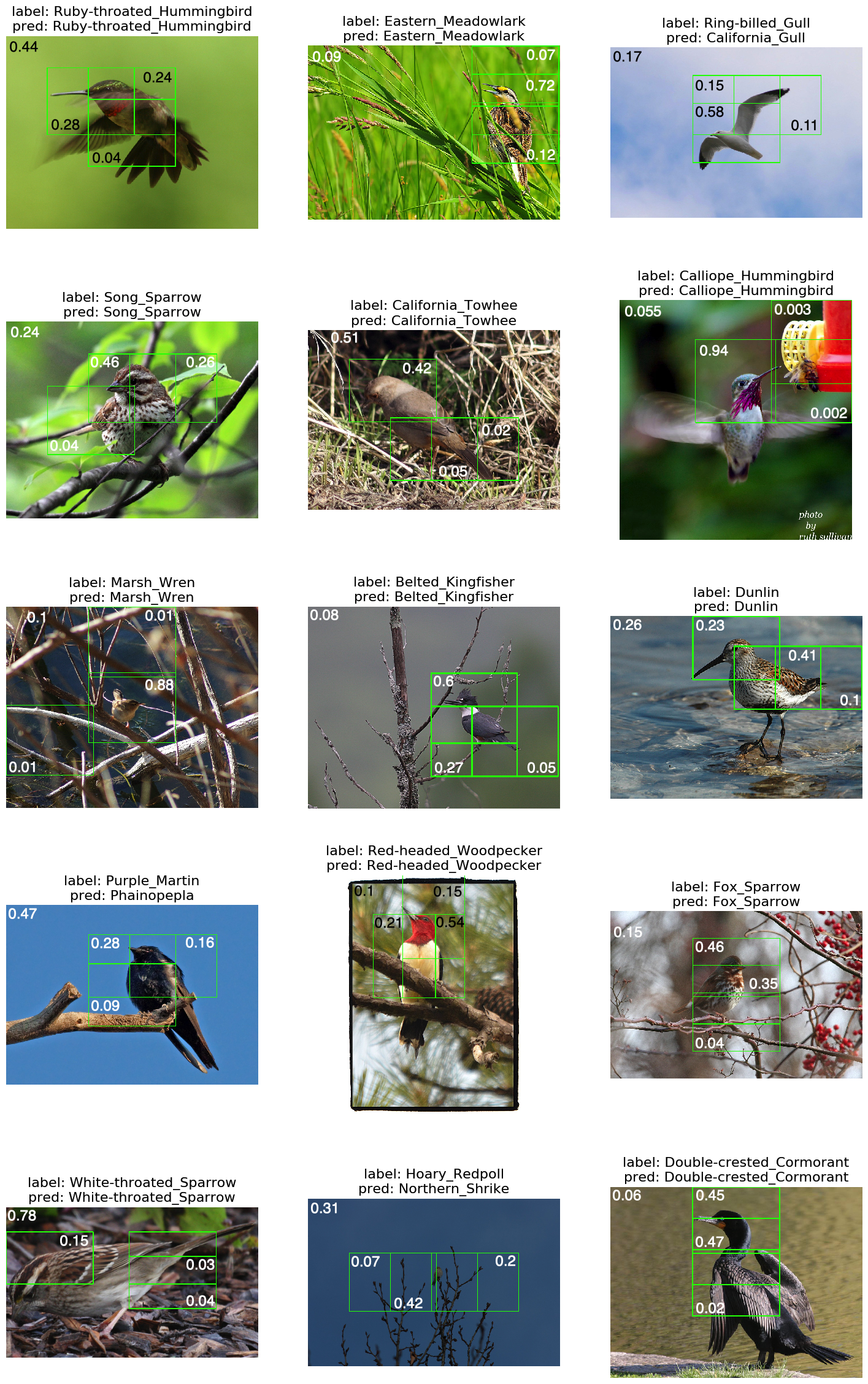}
\caption{Examples from attention policy learned on NABirds. Numeric annotations correspond to weights predicted by the feature weighting module, for the top $3$ locations and the downscaled version of the whole image ($1$st processing level).
Weights sum up to $1$.
For every image, the correct and predicted labels are provided.
}
\label{App_fig_4}
\end{figure*}

In Figure \ref{App_fig_1} we provide examples of the attention policy on the ImageNet validation set with $3$ locations.
In Figure \ref{App_fig_2} we provide examples of the attention policy on the fMoW test set with $2$ locations at the $2$nd processing level.
In Figures \ref{App_fig_3} and \ref{App_fig_4} we provide attention policy examples on the validation sets of CUB-$200$-$2011$ and NABirds respectively. In both cases, $3$ location are attended, and the weights estimated by the feature weighting module are provided as well.

\end{document}